\documentclass[10pt,twocolumn,letterpaper]{article}

\usepackage[pagenumbers]{cvpr} 

\usepackage[dvipsnames]{xcolor}

\definecolor{cvprblue}{rgb}{0.21,0.49,0.74}
\usepackage{comment}
\usepackage[algoruled,linesnumbered,lined,boxed, noend]{algorithm2e}
\usepackage{graphicx}
\usepackage{tabularx}
\usepackage{xcolor}
\usepackage{mathtools}
\usepackage[pagebackref,breaklinks,colorlinks,citecolor=cvprblue]{hyperref}

\def\paperID{2574} 
\def\confName{CVPR}
\def\confYear{2024}

\title{RandMSAugment: A Mixed-Sample Augmentation for Limited-Data Scenarios}

\author{Swarna K Ravindran\\
Duke University\\
Duke Vision Lab, Department of Computer Science\\
{\tt\small swarnakr@cs.duke.edu}
\and
Carlo Tomasi\\
Duke University\\
Duke Vision Lab, Department of Computer Science\\
{\tt\small tomasi@cs.duke.edu}
}

\begin{document}
\maketitle
\begin{abstract}
The high costs of annotating large datasets suggests a need for effectively training CNNs with limited data, and data augmentation is a promising direction. We study foundational augmentation techniques, including Mixed Sample Data Augmentations (MSDAs) and a no-parameter variant of RandAugment termed Preset-RandAugment, in the fully supervised scenario. We observe that Preset-RandAugment excels in limited-data contexts while MSDAs are moderately effective. 
We show that low-level feature transforms play a pivotal role in this performance difference, 
postulate a new property of augmentations related to their data efficiency, and 
propose new ways to measure the diversity and realism of augmentations.
 Building on these insights, we introduce a novel augmentation technique called RandMSAugment that integrates complementary strengths of existing methods. 
RandMSAugment significantly outperforms the competition on CIFAR-100, STL-10, and Tiny-Imagenet. With very small training sets (4, 25, 100 samples/class), RandMSAugment achieves compelling performance gains between 4.1\% and 6.75\%. Even with more training data (500 samples/class) we improve performance by 1.03\% to 2.47\%.
RandMSAugment does not require hyperparameter tuning, extra validation data, or cumbersome optimizations. 
\end{abstract}    
\section{Introduction}
\label{sec:intro}
The success of deep convolutional neural networks (CNNs) is partly due to the use of large manually labelled datasets~\cite{krizhevsky2009learning}. The annotation cost for these datasets is often prohibitive, and training CNNs on \emph{limited data in a fully supervised} setting therefore remains crucial. 

Data augmentation adds new samples to a training set by artificial perturbations of the existing ones. It is especially useful in improving generalization in limited-data scenarios.
Traditional augmentation techniques were based on photometric and geometric transformations, but required extensive hyperparameter tuning \cite{krizhevsky2012imagenet, zagoruyko2016wide}.
More recent approaches \cite{Cubuk_2019_CVPR, lim2019fast, ho2019population, cubuk2020randaugment} automatically tune hyperparameters for the above low-level transforms using reinforcement learning. However, they require validation of the parameters over thousands of labelled samples, which are often not available.  
Although RandAugment \cite{cubuk2020randaugment} reduces the number of parameters to tune, it does not eliminate the data-intensive search for optimal parameters.  Semi-supervised learning (SSL) methods \cite{sohn2020fixmatch, xie2020unsupervised} that also work with limited labels develop a data-efficient version of RandAugment that does not require any parameter input, where the parameters are either set to intuitive fixed values or randomly chosen from given ranges. While the augmented samples may not be the optimal ones for model training, this approach is fast, requires no extra data for tuning, and the resulting model benefits from  the increased diversity of augmented samples. We refer to this version of augmentation as Preset-RandAugment. 

\begin{table}
  \centering
  \scriptsize
  \begin{tabular}{@{}lccc@{}}
    \toprule
    Dataset & Augmentation &  \multicolumn{2}{c}{Samples / Class} \\
     &&  4 & 500 \\
    \midrule
    \multirow{3}{*}{CIFAR-100} & ResizeMix & 13.84\tiny{$\pm$0.18} & \underline{85.25\tiny{$\pm$0.19}}\\
& GMix & \underline{20.02\tiny{$\pm$0.11}} & 84.24\tiny{$\pm$0.12}\\
& { RandMSAugment} & \bf{24.27\tiny{$\pm$0.26}} & \bf{86.28\tiny{$\pm$0.24}}\\
\hline 
\multirow{3}{*}{STL-10} & {\smaller Preset-RandAugment} & \underline{39.77\tiny{$\pm$0.17}} & 91.21\tiny{$\pm$0.13}\\
& HMix & 30.43\tiny{$\pm$0.39} & \underline{91.58\tiny{$\pm$0.09}}\\
&  { RandMSAugment} & \bf{43.50\tiny{$\pm$0.35}} & \bf{94.05\tiny{$\pm$0.05}}\\
\hline 
\multirow{3}{*}{Tiny-Imagenet}  & {\smaller Preset-RandAugment} & \underline{10.54\tiny{$\pm$0.13}} & 67.47\tiny{$\pm$0.10}\\
& ResizeMix & 7.80\tiny{$\pm$0.10} & \underline{67.77\tiny{$\pm$0.22}}\\
& RandMSAugment & \bf{14.48\tiny{$\pm$0.27}} & \bf{69.96\tiny{$\pm$0.27}}\\
    \bottomrule
  \end{tabular}
  \caption{Our method outperforms the state-of-the-art methods Preset-RandAugment\cite{sohn2020fixmatch}, ResizeMix\cite{qin2020resizemix}, GMix and HMix \cite{park2022unified} on CIFAR-100 \cite{krizhevsky2009learning}, STL-10 \cite{coates2011analysis}, and Tiny-Imagenet \cite{chrabaszcz2017downsampled}. Our method (bold) improves over the next best (underlined) by {\bf 4.1\% to 5.24\% } for limited data (4 per class) and by {\bf 1.03\% to 2.47\% } points for abundant data (500 per class).}
  \label{tab:summary}
\end{table}

Mixed Sample Data Augmentations (MSDA) also require no data-dependent hyperparameter tuning. They mix a pair of samples and their labels in a linear combination (Mixup \cite{zhang2017mixup}, Manifold Mixup \cite{verma2019manifold}) or in a cut-and-paste manner (Cutmix \cite{yun2019cutmix}, PuzzleMix \cite{kim2020puzzle}, ResizeMix \cite{qin2020resizemix}). 
 In these methods, the images as well as associated mixing parameters are random, so they require no parameter tuning and lead to greater diversity of the mixed samples. The mixed samples can be seen as  adding noise to the training data, which has a strong regularizing effect \cite{zhang2020does, park2022unified}. 
 These samples along with their soft, non-binary labels help decrease overfitting, even for high-capacity neural networks \cite{zhang2020does, pinto2022using}. The mixed samples are continuous and lie on the boundaries between classes, leading to smoother boundaries that overfit less \cite{oh2023provable, chidambaram2021towards}.

We found that MSDAs are not as effective when there are only, say, 4-25 training samples per class, while Preset-RandAugment has the best performance among existing methods in these scenarios. Even so, there is significant room for improvement, as we show in this paper.

Expanding on prior work~\cite{cubuk2021tradeoffs, geiping2022much}, we seek augmentation samples that (1) have \emph{high diversity}, so that they generate  diverse data variations of the original data samples; (2) are \emph{realistic}, in that they come from the true data distribution; and (3) encourage \emph{faster convergence} by helping the model learn stable and invariant low-level features.
We explain the effectiveness of Preset-RandAugment in limited sample scenarios in terms of these properties and we identify low-level feature transforms as a key contributor to performance. 

Based on these insights, we propose a novel augmentation technique called \emph{RandMSAugment} that integrates complementary strengths: Low-level feature transforms from Preset-RandAugment and interpolation and cut-and-paste from MSDA. We improve diversity through added stochasticity in the mixing process. 
RandMSAugment significantly outperforms the competition (Table ~\ref{tab:summary}) in the fully supervised scenario. It does not require hyperparameter tuning, extra validation data, or cumbersome optimizations. 

We demonstrate superior performance on 3 supervised image classification datasets: CIFAR-100 \cite{krizhevsky2009learning}, STL-10 \cite{coates2011analysis},  and Tiny-Imagenet \cite{chrabaszcz2017downsampled}. RandMSAugment performs the best with more dramatic gains in performance for sparse samples (4, 25, 100 samples / class), ranging from 4.1\% to 6.75\% points. We also improve performance when the samples are more abundant (500 samples / class), with more modest gains ranging from 1.03 \% to 2.47 \% points. 

\section{Related Work}
\label{sec:related_work}

Mixup~\cite{zhang2017mixup} improves generalization by blending pairs of random images and their labels using convex linear combinations. Manifold Mixup~\cite{verma2019manifold} and MoEx~\cite{li2021feature} improve performance further by similarly combining activations in the hidden layers of CNNs as well. 

Cut-and-paste methods like Cutmix~\cite{yun2019cutmix} overwrite a random rectangle from one image with an equal-sized rectangle from another.  This focuses the model's attention on object parts at different locations~\cite{devries2017improved, zhong2020random} and leads it to explore a broader feature space.

\paragraph{Saliency based methods}
The random regions selected by cut-and-paste methods do not always contain salient information, and the mixed labels may be mismatched as a result. Saliency-based methods compute a saliency map for every image in a training batch and rearrange image parts in the mixed samples so as to maximize saliency ~\cite{kim2020puzzle, huang2021snapmix, uddin2020saliencymix, kim2021co}. This reduces label mismatches but requires cumbersome optimizations and large amounts of data~\cite{kang2023guidedmixup, zhu2020automix, dabouei2021supermix, cheung2022transformmix}.  Importantly, the saliency map is computed with a network trained on the \emph{full} Imagenet dataset~\cite{simonyan2013deep}. Such a network is not available in the scarce-samples scenario. Simpler saliency models do not perform as well~\cite{kang2023guidedmixup}.

ResizeMix~\cite{qin2020resizemix} strikes a better balance between accuracy and efficiency: It resizes an entire image before pasting it in a random location in the other, thus preserving object information even if saliency is not maximized.

\paragraph{Generalized MSDA} 
FMix~\cite{harris2020fmix} builds on Cutmix, creating complex-shaped mixing masks through thresholded inverse Fourier transforms of filtered random Gaussian arrays.
 Gridmix~\cite{baek2021gridmix} randomly blends equal-sized grid cells from two images to create mixed samples.
Automix~\citep{zhu2020automix} uses a neural mixing module for automatically generating Mixup-like masks, training it alongside the image classifier with a momentum-based method for stabilizing the bi-level optimization.
Park \etal~\cite{park2022unified} generalize MSDAs with two methods called HMix and GMix. HMix first creates a CutMix-like sample by cut-and-paste and then interpolates both samples outside the box \textit{\`{a} la} Mixup. GMix smooths the boundaries of Cutmix boxes by replacing the binary mask with a feathered one. The paper also studies the regularization effects of MSDA methods theoretically.

\paragraph{Related to RandAugment}
 CTAugment~\cite{berthelot2019remixmatch} starts like Preset-RandAugment but adjusts transformation magnitudes during training, based on how closely the model's prediction aligns with the true label. Although it involves extra computations, its performance is comparable to Preset-RandAugment~\cite{sohn2020fixmatch}.
AugMix~\cite{hendrycks2019augmix} makes an augmented sample by linearly combining three images obtained by random sequences of Preset-RandAugment low-level transformations. The training loss is a combination of the standard cross-entropy loss on the original images and a consistency loss between the original and augmented image. 

\section{Properties of Augmentations}
\label{sec:properties}
\begin{figure}
  \centering
    \includegraphics[width=0.5\linewidth]{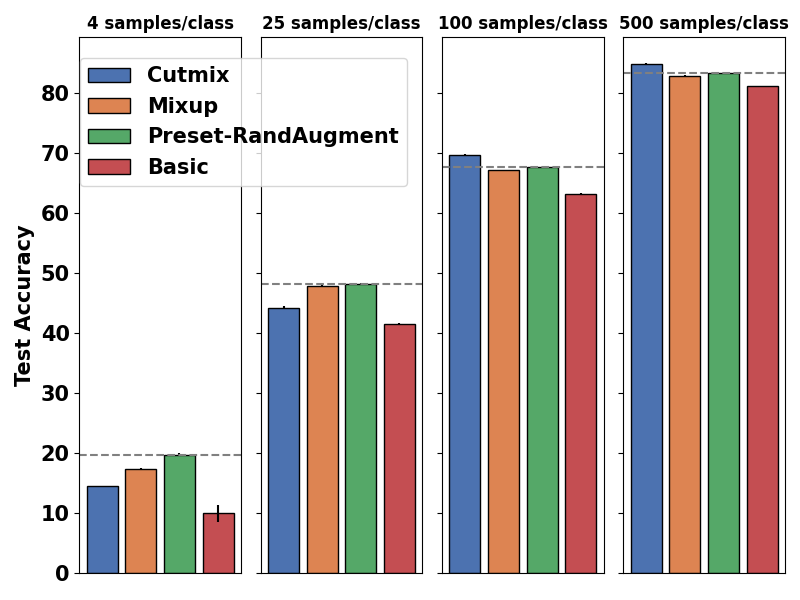}
    \caption{Performance of SotA augmentation methods on CIFAR 100~\cite{krizhevsky2009learning} trained with WideResnet-28-10~\cite{zagoruyko2016wide} for different number of training samples per class. Preset-RandAugment (green) performs better than MSDA (Cutmix\cite{yun2019cutmix} and Mixup\cite{zhang2017mixup}) with few training samples whereas the reverse is true with plentiful samples. As expected, the Basic augmentation that horizontally flips and crops the image has the lowest performance throughout. }
    \label{fig:acc}
  \end{figure}

 We compare the performance of foundational augmentation techniques in Figure~\ref{fig:acc} and note that 
Preset-RandAugment has the best performance  in low data regimes, with just 4-25 training samples per class. MSDA methods are not as effective with limited data as they are when data is abundant. 

It is widely recognized that augmentations need to possess two crucial attributes: high diversity and realism \cite{cubuk2021tradeoffs}, so that the augmented data represents a wide range of possible and plausible real-world scenarios. 

Additionally, suitable inductive bias can yield faster model convergence and better generalization by restricting the model's hypothesis space \cite{goodfellow2016deep}.
In limited data scenarios it is best to focus this bias on less class-specific patterns such as edges and textures, that is, to induce the model to learn stable low-level features in the early layers of a network.
We formalize all the above properties below.


\subsection{Definitions}
\label{sec:definitions}
Supervised training of a predictor $f \in F$ minimizes the risk (average loss) \( R(f) = \mathbb{E}_{(\mathbf{x}, \mathbf{y}) \sim P} [l(f(\mathbf{x}), \mathbf{y})]\), over the data distribution \(P(\mathbf{x}, \mathbf{y})\) of the image and target class pairs. 
\(P(\mathbf{x}, \mathbf{y})\) is unknown, but a training dataset \(\mathcal{D}_e = \{(\mathbf{x}_i,\mathbf{y}_i)\}_{i=1}^m\) that is randomly sampled from $P$ is available.  \(P(\mathbf{x}, \mathbf{y})\) can thus be approximated by an empirical distribution \( P_e(\mathbf{x}, \mathbf{y})\) that places $m$ delta functions at \((\mathbf{x}_i, \mathbf{y}_i)\): \( P_e(\mathbf{x}, \mathbf{y}) = \frac{1}{m} \sum_{i=1}^{m} \delta_{\mathbf{x}_i} (\mathbf{x}) \delta_{\mathbf{y}_i} (\mathbf{y}). \) 

The quality of the approximation in $P_e$ is poor  when $\mathcal{D}_e$ is either sparse or only covers a small part of the support of $P$.
We can then estimate $P$ better using a \emph{vicinal distribution} \cite{chapelle2000vicinal} \( P_a( \tilde{\mathbf{x}}, \tilde{\mathbf{y}} | \mathbf{x}_i,\mathbf{y}_i )\) that models the space of all augmented samples $(\tilde{\mathbf{x}},\tilde{\mathbf{y}})$
in the vicinity of \((\mathbf{x}_i, \mathbf{y}_i)\).
A good augmentation generates \( (\tilde{\mathbf{x}}, \tilde{\mathbf{y}}) \) that lie within the convex hull of the samples \( ({\mathbf{x}}, {\mathbf{y}}) \sim \mathcal{D}_e\).
Thus, $P$ can be approximated better as:
\( P_a(\tilde{\mathbf{x}}, \tilde{\mathbf{y}}) = \sum_{i=1}^{m} P_a( \tilde{\mathbf{x}}, \tilde{\mathbf{y}} | \mathbf{x}_i,\mathbf{y}_i ) P( \mathbf{x}_i,\mathbf{y}_i )\) \cite{chapelle2000vicinal}, and
the augmented dataset \(\mathcal{D}_a = \{(\tilde{\mathbf{x}}_i,\tilde{\mathbf{y}}_i)\}_{i=1}^n\) can be viewed as samples from $P_a$ \cite{zhang2017mixup}. Training then minimizes the vicinal risk
\( R_{v}(f) = \mathbb{E}_{(\tilde{\mathbf{x}}, \tilde{\mathbf{y}}) \sim P_a} [l(f(\tilde{\mathbf{x}}), \tilde{\mathbf{y}})]\).

The following properties of augmentation methods are valuable when $\mathcal{D}_e$ is very small:

\noindent\textbf{\textit{1.}} \emph{High diversity} is the average spread or variance of the vicinal distribution $P_a( \tilde{\mathbf{x}}, \tilde{\mathbf{y}} | \mathbf{x}_i,\mathbf{y}_i)$ over all \((\mathbf{x}_i,\mathbf{y}_i) \in \mathcal{D}_e\). A wide spread ensures a denser coverage of $P$ around each original sample \(({\mathbf{x}}_i, {\mathbf{y}}_i)\). 
    
\noindent\textbf{\textit{2.}} \emph{High similarity of $P_a$ to $P$}: The augmented samples \( (\tilde{\mathbf{x}}, \tilde{\mathbf{y}}) \) must be realistic and come from the true data distribution $P$. Since $P$ is unknown and approximated by $P_e$, we measure realism in $P_a$ by its similarity to $P_e$ as described in Section~\ref{sec:similarity}. 
    Setting $P_a=P_e$ would achieve maximum realism in $P_a$ but at the expense of diversity (property 1), so we must strike a balance between these two properties.
    
\noindent\textbf{\textit{3.}} \emph{Fast convergence through Low-level Feature Invariance}: The neural function $f$ extracts low-level and high-level features from $\mathbf{x}$ in its early and later layers respectively:  $f({\mathbf{x}}) = (f^l({\mathbf{x}}), f^h({\mathbf{x}}))$. Augmented samples $(\tilde{\mathbf{x}}, \tilde{\mathbf{y}}) \sim P_a$ that are stable across similar ${\tilde{\mathbf{x}}}$ result in faster convergence to a better solution $f^*$. Stable $f^l({\tilde{\mathbf{x}}})$ are easier to achieve when training samples are sparse, since low-level feature information is class-agnostic and widely prevalent. It is harder to directly achieve stable $f^h({\tilde{\mathbf{x}}})$ with sparse samples. 
    

 In order to measure all the above properties, we sample $k$ times from \( P_a( \tilde{\mathbf{x}}, \tilde{\mathbf{y}} | \mathbf{x}_i,\mathbf{y}_i )\) around a data sample \((\mathbf{x}_i, \mathbf{y}_i) \in \mathcal{D}_e \) to produce an augmentation cloud \( \prescript{k}{}{\mathcal{D}}_a^i \).
In MSDA methods, \( P_a( \tilde{\mathbf{x}}, \tilde{\mathbf{y}} | \mathbf{x}_i,\mathbf{y}_i )\) depends on \((\mathbf{x}_i, \mathbf{y}_i)\) as well another sample \((\mathbf{x}_j, \mathbf{y}_j)\) drawn from $\mathcal{D}_e$.
Further, $\mathcal{D}_e$ is split into training, validation and test datasets 
$\mathcal{T}_e$, $\mathcal{V}_e$ and $\mathcal{S}_e$.
We analyze foundational augmentation methods in terms of these properties next and use $\mathcal{T}_e$, $\mathcal{V}_e$ from the CIFAR-100 dataset for this analysis. 
In all experiments, averages and errors are calculated across 3 distinctly seeded models.
\subsection{Diversity}

\begin{table}
    \centering
    \scriptsize
    \begin{tabular}{|c|c|}
    
          \hline
            Method & Diversity \\
            \hline \hline
            Preset-RandAugment & {\bf{2516.78}}\\
           Cutmix & 1775.41\\
           Mixup & 1643.86\\
           Basic & 1141.87\\
             \hline
        \end{tabular}
  \caption{Diversity $ \langle{\sigma_a^{2}} \rangle$ for different foundational augmentation methods on the CIFAR-100 dataset. Preset-RandAugment has the highest diversity. }
    \label{tab:diversity}
  \end{table}

\begin{figure}[t]
\centering
\begin{tabular}{@{}c@{}c@{}c@{}}
 \renewcommand{\arraystretch}{0.7}
\includegraphics[scale=0.165]{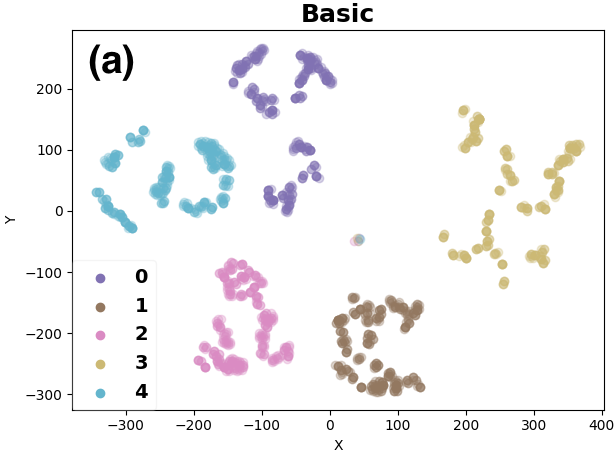} &
\includegraphics[scale=0.165]{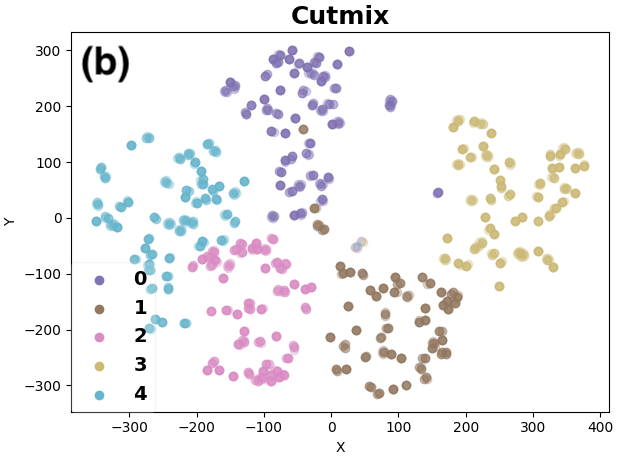} &
\includegraphics[scale=0.165]{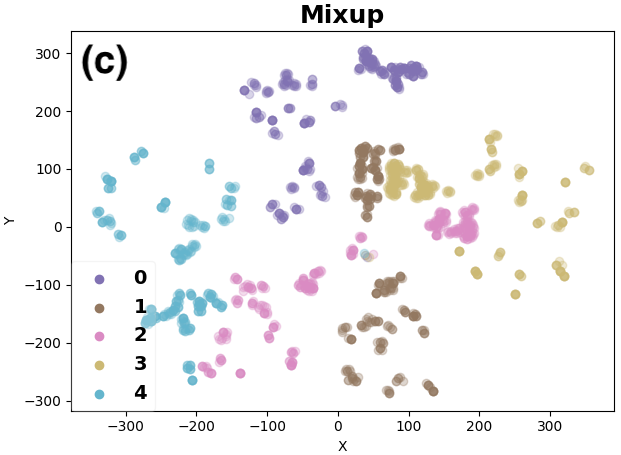}\\
\includegraphics[scale=0.165]{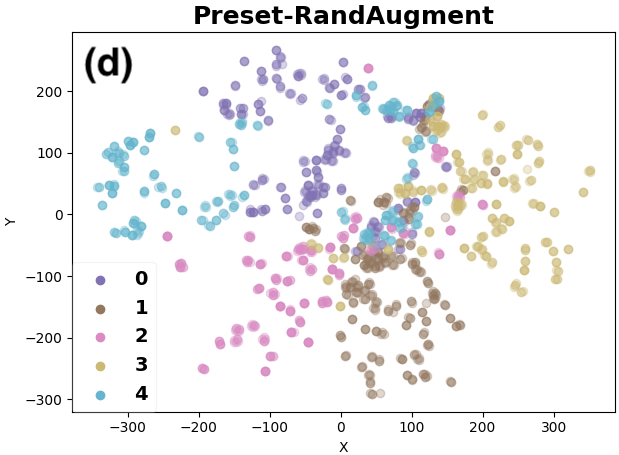} & 
\includegraphics[scale=0.165]{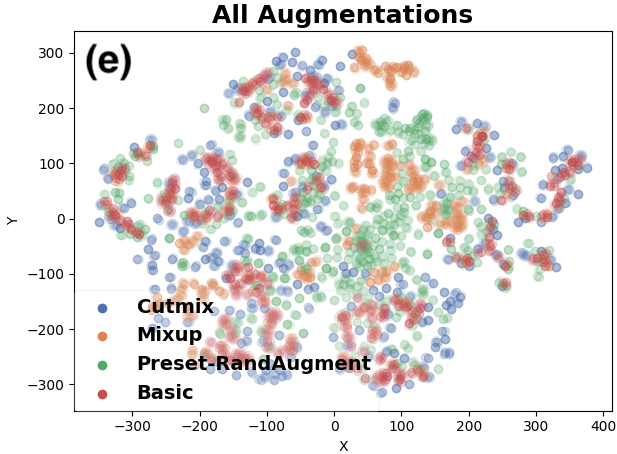} \\
\end{tabular}
\caption{ (a-d) Visualizing \(\prescript{10,000}{}{\mathcal{D}}_a^i\)  in 2D using t-SNE for 5 randomly drawn samples $x_i \in \mathcal{T}_e$ and for different augmentation methods. 
Plot (e) combines the clouds for all methods, with different colors for different methods. Preset-RandAugment's  (green) is the most  dispersed in the embedding space. } 
\label{fig:diversity}
\end{figure}

All the foundational augmentation methods -- Preset-RandAugment, Cutmix and Mixup -- generate \emph{dynamic} augmentations by generating  augmentation parameters randomly in each training iteration.  This stochasticity increases diversity and improves generalization over simple augmentations like flips and crops \cite{Cubuk_2019_CVPR}.
 
Cubuk \etal \cite{cubuk2021tradeoffs} measure the diversity of an augmentation method using the training risk (at convergence) of a model trained with it, based on the intuition that it is harder to overfit with more diverse data. However, the magnitude of the risk depends on other factors as well, such as the capacity of the model, size of the training set, optimization settings \etc Also, \cite{marcu2022effects} found the claims made by \cite{cubuk2021tradeoffs} not to hold for more recent augmentation methods. 

Instead, we measure diversity as the
average variance of the vicinal distribution  \( P_a( \tilde{\mathbf{x}}, \tilde{\mathbf{y}} | \mathbf{x}_i,\mathbf{y}_i )\)  around a data sample \((\mathbf{x}_i, \mathbf{y}_i) \in \mathcal{T}_e\). We compute the augmentation cloud \( \prescript{k}{}{\mathcal{D}}_a^i \) with $k=10,000$, reduce it to 100 dimensions with PCA \cite{wold1987principal}, and compute the variance $(\sigma_a^{2})^i$ of the reduced point cloud. 
Diversity, denoted by $ \langle{\sigma_a^{2}} \rangle$, is the average \((\sigma_a^2)^i\) over all \((\mathbf{x}_i, \mathbf{y}_i) \in  \mathcal{T}_e\), shown for different methods in Table~\ref{tab:diversity}. We measure diversity around a single image sample for Preset-RandAugment, and a pair of images for MSDA.
Figure~\ref{fig:diversity} (a-e) visualizes the point clouds for some random samples from $\mathcal{T}_e$ in 2D using t-SNE embeddings~\cite{van2008visualizing}. 

Preset-RandAugment's augmentations has higher diversity (Table~\ref{tab:diversity}) compared to MSDA.
It uses photometric operations like equalize, contrast, color and brightness adjustments that redistribute and shift pixel intensities  (Figure~\ref{fig:diversity}(d-e)) and increase model invariance to real-world lighting variations.
The next section shows that despite this higher diversity, Preset-RandAugment's outputs are realistic and close to the true distribution $P$.

\subsection{Similarity to True Distribution}
\label{sec:similarity}

An augmentation method must strike a balance between producing diverse samples and ensuring that those samples are still realistic and relevant. 
A sample in $P_a$ is realistic when it comes from the true distribution $P$, which $P_e$ approximates.
So we measure the similarity of $P_a$ to $P_e$.

Several strategies are used in literature to measure distributional similarity  \cite{lim2019fast, cubuk2021tradeoffs, facciolo2017automatic}.
Facciolo \etal \cite{facciolo2017automatic} train unsupervised variational auto-encoders (VAEs) on the real and the augmented data and measure the similarity between the learned representations by mutual information. Training a VAE is cumbersome and the resulting measure relies heavily on the VAE generating good quality samples. 

Other methods \cite{lim2019fast, cubuk2021tradeoffs}
train a model on the clean training data $\mathcal{T}_e$, and  measure its accuracy on an augmented validation set 
\( \prescript{1}{}{\mathcal{D}}_a^{\mathcal{V}} = \bigcup_{i} \{\prescript{1}{}{\mathcal{D}}_a^i: (\mathbf{x}_i, \mathbf{y}_i) \in \mathcal{V}_e \} \),   the aggregate of all the individual augmentation clouds for each point in $\mathcal{V}_e$.
The hypothesis is that if the pattern of the data \( \prescript{1}{}{\mathcal{D}}_a^{\mathcal{V}} \) generated by a given method follows that of $\mathcal{V}_e$, then their validation accuracies will be close.

In contrast, we use the cross entropy score instead of accuracy. Accuracy computation requires a single predicted label for an input. However, MSDA makes soft labels for pairs of mixed images. Computing accuracy requires thresholding these soft labels and can produce incorrect results, making cross entropy more appropriate in this case.   

\begin{figure}[t]
\centering
 \renewcommand{\arraystretch}{0.7}
\begin{tabular}{cc}
\includegraphics[scale=0.18]{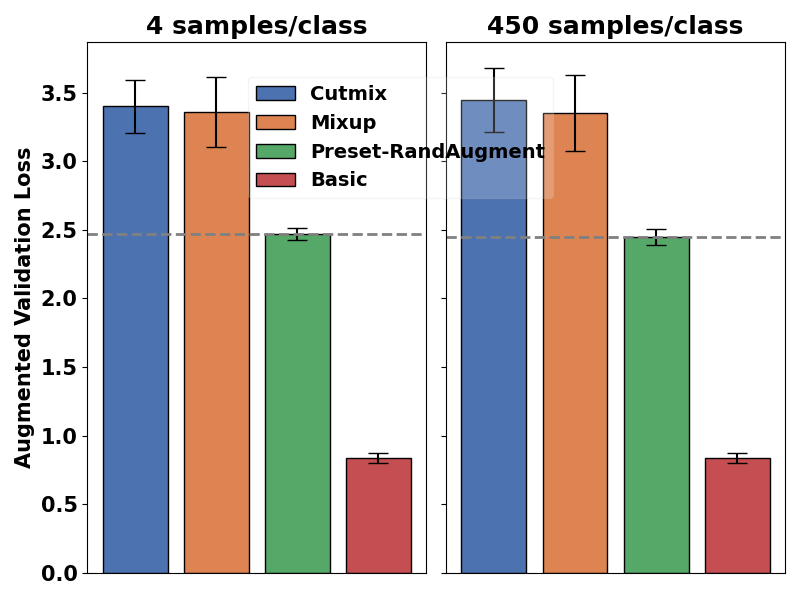} &
\includegraphics[scale=0.18]{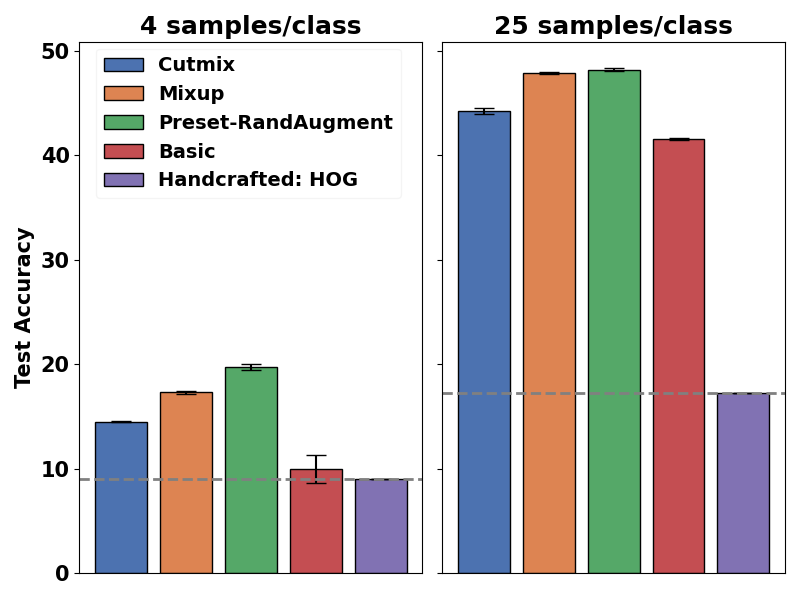}\\
\smaller (a) & \smaller (b) \\
\end{tabular}
\caption{(a) Distributional similarity of samples from different augmentation methods to $P_e$, measured by the cross entropy risk~($\downarrow$)  on an augmented validation set \( \prescript{1}{}{\mathcal{D}}_a^{\mathcal{V}} \).  (b) Comparing a simple classifier learned with handcrafted features with foundational augmentation methods trained with CNNs. }
\label{fig:similarity_handcraft}
\end{figure}

We measure distributional similarity of an augmentation distribution $P_a$ to the empirical distribution $P_e$ by the cross entropy risk  \( R(f^*) = \mathbb{E}_{(\tilde{\mathbf{x}}, \tilde{\mathbf{y}}) \sim \prescript{1}{}{\mathcal{D}}_a^{\mathcal{V}}} [l(f^*(\tilde{\mathbf{x}}), \tilde{\mathbf{y}})]\) of a model $f^*$ trained on the clean $\mathcal{T}_e$, when it has sparse and abundant data. 
Figure~\ref{fig:similarity_handcraft}(a) shows that 
Basic Augmentation (flips and crops) samples have the lowest risk, and are closest to $P_e$, followed by Preset-RandAugment. On the other hand, Cutmix and Mixup samples are the farthest from $P_e$. 

\begin{figure}[t]
\centering
\scriptsize
\begin{tabular}{@{}ccccc@{}}
\includegraphics[scale=0.6]{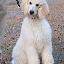} & 
\includegraphics[scale=0.6]{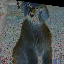} & 
\includegraphics[scale=0.6]{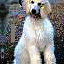} & 
\includegraphics[scale=0.6]{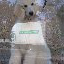} &
\includegraphics[scale=0.6]{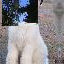} \\

\includegraphics[scale=0.6]{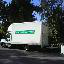} & 
\includegraphics[scale=0.6]{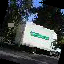} & 
\includegraphics[scale=0.6]{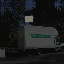} & 
\includegraphics[scale=0.6]{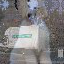} &
\includegraphics[scale=0.6]{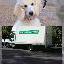} \\

(a) Input &  \multicolumn{2}{c}{(b) Preset-RandAugment} & (c) Mixup & (d) Cutmix \\

\end{tabular}
\caption{For an example from the poodle and van classes, Preset-RandAugment samples model real-world changes like rotations and lighting variations, and preserve high-level object information. On the other hand, MSDA samples dilute semantic information and appear unrealistic.}
\label{fig:realism}
\end{figure}

In contrast to MSDA samples, Preset-RandAugment samples preserve the high-level semantics of the object, causing the model to perceive them to be more realistic.  Further, Preset-RandAugment generates augmentations insensitive to changes in lighting and viewpoint.  Figure~\ref{fig:realism} shows examples from both types of methods.

\subsection{Low-Level Feature Invariance}

 \begin{figure*}[t]
\centering
\begin{tabular}{@{}c@{}c@{}c@{}c@{}}
\includegraphics[scale=0.175]{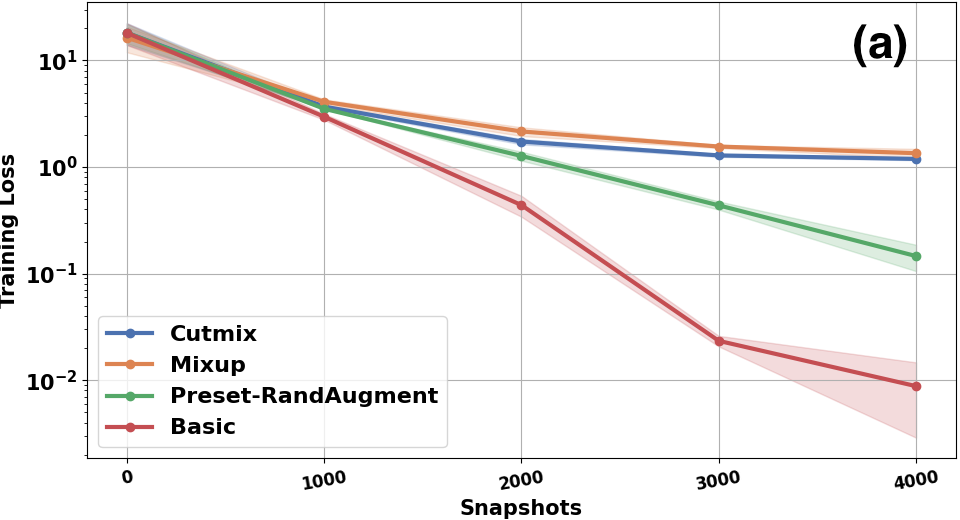} &
\includegraphics[scale=0.175]{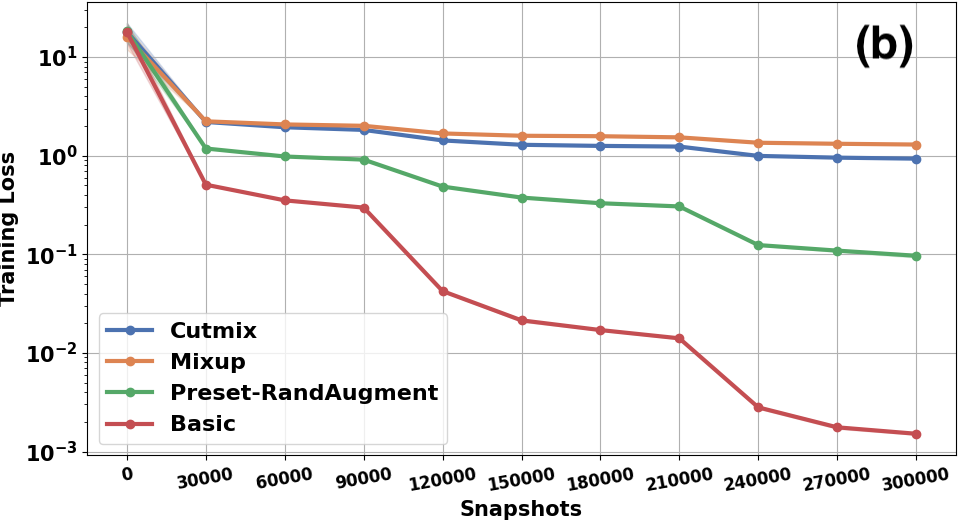}  &
\includegraphics[scale=0.175]{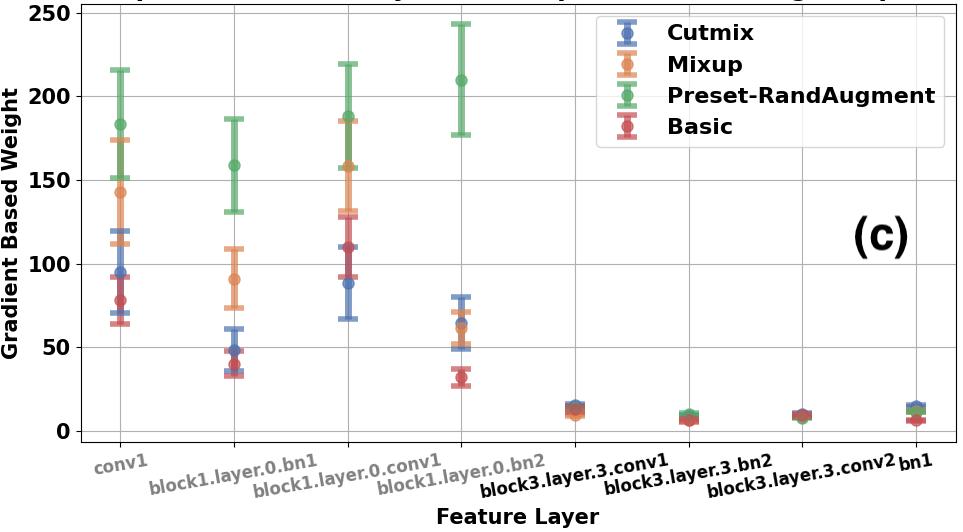} &
\includegraphics[scale=0.175]{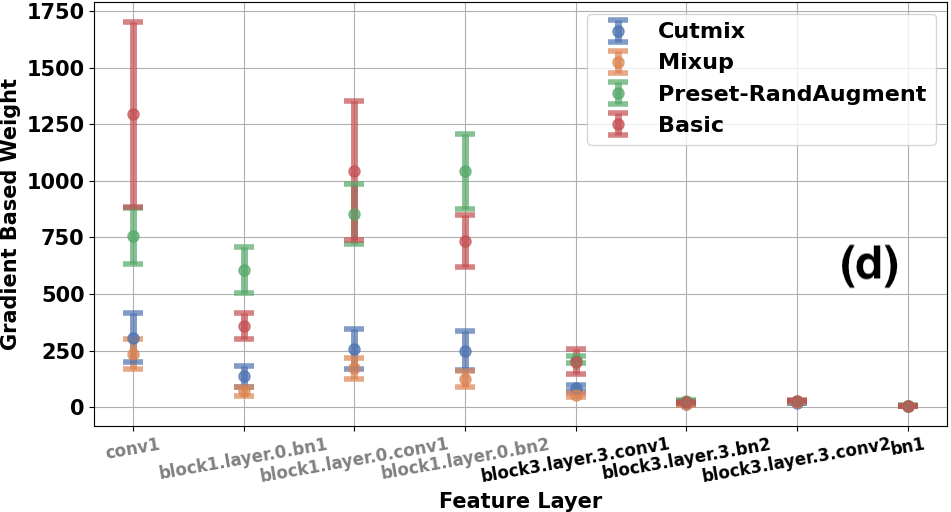} \\
\end{tabular}

\caption{(a-b) Learning curve for different augmentation methods, and for (a) sparse (4 per class) and (b) abundant (450 per class)  training samples. (c-d) Gradient based weights $w_r^c$ averaged over all  samples  $(\mathbf{x},\mathbf{y})  \in \mathcal{V}_e$, and over 3 differently seeded models is shown for 4  lowest (gray) and highest (black) feature maps in the model. The model in (c) is trained with sparse (4 per class) samples and in (d) it is trained with abundant (450 per class)  samples.}
\label{fig:lc_gradcam}
\end{figure*}

Augmentation samples ${\tilde{\mathbf{x}}}$  that encourage the model to learn invariant features $f({\tilde{\mathbf{x}}})$ 
lead to faster convergence and better generalization \cite{goodfellow2016deep}.
It is easier to achieve low-level feature $f^l({\tilde{\mathbf{x}}})$  stability due to the ubiquity  of low-level information in sparse data settings, as detailed in Section~\ref{sec:definitions}. 

The convergence rate is measured from the plots of risk $R(f)$ versus training iteration, shown in Figure~\ref{fig:lc_gradcam}(a-b). The model $f$ is trained on \( \prescript{k}{}{\mathcal{D}}_a^{\mathcal{T}} = \bigcup_{i} \{\prescript{k}{}{\mathcal{D}}_a^i: (\mathbf{x}_i, \mathbf{y}_i) \in \mathcal{T}_e \}\).
Preset-RandAugment models converge consistently faster than MSDAs with both limited (4 per class) and plentiful (450 per class) training data. 
Preset-RandAugment is based on low-level feature manipulations that lead to model invariance to such changes. In contrast, the blending operations and sharp discontinuities in MSDA methods 
dilute low-level and high-level semantic information (Figure~\ref{fig:realism}). This makes it harder for the network to learn stable features.

\paragraph{Promoting Convergence Improves Generalization} 
We show that improving the training speed of the MSDA method, Cutmix, results in better generalization. The Basic augmentations enable fast model convergence (Figure~\ref{fig:lc_gradcam}(a-b)). 
 We inject these augmentations during the first 10,000 iterations of training and then present Cutmix samples for the remainder. 
This process is motivated by curriculum learning \cite{bengio2009curriculum}, and shows that promoting convergence even during just the early part of training can improve the generalization of Cutmix. See Table~\ref{tab:early}. 


\begin{table}
    \centering
    \scriptsize
    \begin{tabular}{|c|c|c|}
    \hline
    Aug & Accuracy (\%, $\uparrow$) & Risk at 50K iterations ($\downarrow$)\\
    \hline
        Cutmix & 14.50\tiny{$\pm$0.04 }  & 0.67 \tiny{$\pm$0.21 }\\
Cutmix-Faster & \bf{15.32\tiny{$\pm$0.39 }} &  \bf{0.54 \tiny{$\pm$0.13 }} \\
\hline
    \end{tabular}
    \caption{ For 4 samples/class, introducing Basic augmentations in Cutmix-Faster enables faster convergence (lower risk at 50K iterations) and better accuracy than Cutmix. }
    \label{tab:early}
\end{table}

\paragraph{Impact of Learned Low-level Features on Output Prediction} 

We use  Gradient-weighted Class Activation Mapping (Grad-CAM) \cite{selvaraju2017grad} to show that models that converge faster also learn more impactful low-level features. Grad-CAM is a model interpretability tool that reveals the feature maps in the model that most influence the model's prediction.
It computes the gradient of the model's score for a target class $c$ with respect to the feature map and global-average-pools it to obtain a single weight $w_r^c$ per feature map \( f^r \). This weight measures the sensitivity of the score for class $c$ to changes in the feature map \( f^r \).
We compute $w^c$ using the ground-truth class $\mathbf{y}^c=1$, when the model's maximum prediction aligns with the true class, \ie $\hat{\mathbf{y}}^c = \mathbf{y}^c$. 
Figure~\ref{fig:lc_gradcam}(c-d) shows $w_r^c$ for $r=l$ low-level and $r=h$ high-level feature maps in various scenarios. 

For models trained with Preset-RandAugment on both sparse and plentiful data, 
 low-level features $f^{r=l}$ have a significant influence on the model's correct predictions. With them, Preset-RandAugment model achieves fast convergence (Figure~\ref{fig:lc_gradcam}(a-b)) and generalizes better (Figure~\ref{fig:acc}). 

The Basic augmentation model, which also converges fast, learns good $f^{l}$ when the data is plentiful. However for sparse data, these augmentations have very low diversity (Table~\ref{tab:diversity}) and end up overfitting.

Mixup models learn more impactful $f^{l}$ compared to Cutmix and also generalize better than it. 
It seems that the blended images in Mixup dilute $f^{h}$, preventing the network from relying on them. However, edges and gradients are still present, forcing the network to rely more on $f^{l}$.

\paragraph{Invariant Low-Level Features Aid Generalization} 
Low-level features providing photometric and geometric invariance, such as Scale Invariant Feature Transform (SIFT) \cite{lowe2004sift}, Histogram of Oriented Gradients (HOG) \cite{dalal2005histograms} \etc, were key in the pre-deep learning era for generalization with scarce data.
In Figure~\ref{fig:similarity_handcraft}(b), we compare the foundational augmentations trained on deep CNNs against the HOG + SVM (Support Vector Machines) approach. The manually designed features are almost on par with deep features trained using Basic augmentation when data is very scarce (only 4  per class). This highlights the generalization power of  low-level feature invariance when working with limited data. 
We note that with more data (25 per class) deep features easily overpower handcrafted features.
Preset-RandAugment uses low-level transforms \emph{and} deep features, and excels in the sparse data regime. 


\subsection{Preset-RandAugment Effectiveness}
Thus, the performance of an augmentation method depends on a tradeoff between (1) diversity, (2) realism, (3) invariant low-level feature learning. 
Preset-RandAugment is the most diverse among foundational methods and ranks second in both distributional similarity and convergence rate. Its good balance among all 3 properties makes it particularly effective when training data is limited.

The key contributor to Preset-RandAugment's performance is its use of photometric and geometric transforms. These low-level feature transforms produce augmentations that are diverse, realistic, and promote low-level feature invariance, which speeds up model convergence. 

On the other hand, MSDA samples provide regularization, reduce overfitting, \cite{zhang2020does, pinto2022using, park2022unified} and enable better generalization by 
learning smoother decision boundaries \cite{oh2023provable, chidambaram2021towards}. These properties are complementary to that of Preset-RandAugment and can be combined to produce  better generalization, as we discuss next.

\section{RandMSAugment}
\label{sec:method}
RandMSAugment is a novel augmentation that integrates the complementary, key operations of foundational methods to produce better generalization. We showed earlier the critical role of \emph{low-level feature transforms} in Preset-RandAugment.  We combine this with two key operations used to mix samples in MSDA methods: \emph{linear interpolation}, (Mixup, Manifold Mixup, HMix, GMix \etc) and \emph{cut-and-paste} (Cutmix, ResizeMix \etc). To preserve salient object information in the mixed sample, we choose the efficient and parameter-tuning-free idea of \emph{resize-and-paste} proposed in ResizeMix \cite{qin2020resizemix}, which cuts, resizes, and pastes the entire first image in a random location in the second.

\SetKwProg{Proc}{Procedure}{}{}
\DontPrintSemicolon 
\begin{algorithm}
\smaller
\caption{Components of RandMSAugment}
\label{alg:component}

\KwData{\(\mathbf{x}_1, \mathbf{x}_2 \gets \text{images} \), \(\mathbf{y}_1, \mathbf{y}_2 \gets \text{labels} \)}
\KwResult{\(\mathbf{x}^M_{1,2},\mathbf{y}^M_{1,2}\) }


\SetKwFunction{FResizePaste}{ReszPst}
\SetKwFunction{FResize}{Resize}
\SetKwFunction{FRandomChoice}{Choice}
\SetKwFunction{FInterpolate}{Interp}
\SetKwFunction{FFeatTransform}{FeatXform}
\SetKwFunction{FMixingMask}{MixingMask}
\SetKwFunction{FSequentialApply}{SequentialApply}

\SetKwProg{Fn}{Function}{:}{}
\Fn{\FRandomChoice{$S$, $n$[1]}}{
    $s \gets$ select $n$ elements from $S$ with uniform probability\;
    \Return{$s$}\;
}
\Fn{\FFeatTransform{\(\mathbf{x}, \mathbf{y}, \boldsymbol{\ell}[\text{None}], \boldsymbol{\theta}[\text{None}]\)}}{
   $\mathcal{L} \gets \text{Set of 14 low-level transforms from }$ \cite{cubuk2020randaugment}\; 
    $\Theta \gets \text{Magnitude Range for each } \mathcal{L}$ from \cite{cubuk2020randaugment}\;
    \If{\( \mathbf{l,\theta} \) is not None}{
    $\boldsymbol{\ell}, \boldsymbol{\theta} \gets$ \FRandomChoice{$L$, 2}, \FRandomChoice{$\Theta$, 2}\;
    }
     \(\mathbf{x}^{ft} \gets\) \FSequentialApply{\(\mathbf{x}, \boldsymbol{\ell}, \boldsymbol{\theta}\)}  \;
\Return{\(\mathbf{x}^{ft}, \mathbf{y}, \boldsymbol{\ell}, \boldsymbol{\theta}\)}\;
}

\Fn{\FInterpolate{\(\mathbf{x}_1, \mathbf{x}_2,\mathbf{y}_1, \mathbf{y}_2\)}}{
    $\lambda \gets$ \FRandomChoice{[0:1]}\;
\(\mathbf{x}^{ip}_{1,2}, \mathbf{y}^{ip}_{1,2} = \lambda (\mathbf{x}_1, \mathbf{y}_1) + (1-\lambda) (\mathbf{x}_2, \mathbf{y}_2)\) \;
    \Return{\(\mathbf{x}^{ip}_{1,2}, \mathbf{y}^{ip}_{1,2}\)}\;
}
\Fn{\FResizePaste{\(\mathbf{x}_1, \mathbf{x}_2,\mathbf{y}_1, \mathbf{y}_2, \lambda_s[\text{None}]\)}}{
\If{\( \lambda_{s} \) is not None}{
${\lambda_{s}} \gets $ \FRandomChoice{[0:1]}\;}
 $p, q \gets$ \FRandomChoice{[0: $x_{w}$]}, \FRandomChoice{[0:$x_{h}$]}  \;
 
$\mathcal{M} \gets$ \FMixingMask{$(p,q), (x_{w}, x_{h}), \lambda_{s}$} \;
  \(\mathbf{x}^{rp}_{1,2} = \mathcal{M} \odot\) \FResize{$\mathbf{x}_1, \lambda_{s}$} + \((1 - \mathcal{M}) \odot \mathbf{x}_2\) \;
 \(\mathbf{y}^{rp}_{1,2} = \lambda_{s} \mathbf{y}_1 + (1-\lambda_{s}) \mathbf{y}_2 \) \;
 \Return{\(\mathbf{x}^{rp}_{1,2}, \mathbf{y}^{rp}_{1,2}\)}\;
}

\end{algorithm}

We combine the set of operations, [{\small \FFeatTransform, \FInterpolate, \FResizePaste}] in a way that is inspired by Preset-RandAugment. We choose two operations from the set uniformly at random in every training iteration, and apply them sequentially to the image. 
The operation parameters are also chosen uniformly at random from a predefined range or using hyperparameters recommended by each method's first paper. 
These additional layers of stochasticity in the combination process improve the diversity of the mixed samples. 
 Our method has no new hyperparameters, requires no validation data, and involves no cumbersome optimizations. 

\subsection{Component Functions}

Let $\mathbf{x}_1, \mathbf{x}_2 \in \mathbb{R}^m$ be two images with one-hot vector class label $\mathbf{y}_1, \mathbf{y}_2 \in \mathbb{R}^C$, where $C$ is the number of classes.
In {\small \FFeatTransform}, two operations $\boldsymbol{\ell}$ are  chosen from a set of low level operations $\mathcal{L}$ (rotation, equalizing \etc). Their magnitudes $\boldsymbol{\theta}$ are also randomly chosen from  a range of values $\Theta$, where $\mathcal{L}$ and $\Theta$ are defined by \cite{cubuk2020randaugment}. The operations $\boldsymbol{\ell}$ are applied sequentially to $\mathbf{x}$. These low level transforms do not affect the label $\mathbf{y}$. In {\small \FInterpolate},  mixed samples are generated as the convex combination of two images (and their labels), $\mathbf{x}_1$ ($\mathbf{y}_1$) and $\mathbf{x}_2$ ($\mathbf{y}_2$) in $\mathbb{R}^m$ (and $\mathbb{R}^C$). 
The mixing coefficient \( 0 \leq \lambda \leq 1 \) is randomly sampled from $\text{Beta}(\alpha, \alpha)$, with parameter $\alpha$  specified in \cite{zhang2017mixup}. 
The {\small \FResizePaste} method first computes a binary mask $\mathcal{M}$ at a randomly chosen paste box location $(p,q)$ within the image $\mathbf{x}$ with a relative area $\lambda_{s}$. 
The scale factor \( 0 \leq \lambda_{s} \leq 1 \) is chosen randomly from a range specified in \cite{qin2020resizemix}. {\small \FResizePaste}  resizes $\mathbf{x}_1$ by a factor of $\lambda_{s}$ and then combines it with $\mathbf{x}_2$ using mask $\mathcal{M}$.
The mixed label is obtained as the weighted vector sum of the original labels, with the weights determined by the proportions of the merged areas.
All these component functions are defined in Algorithm~\ref{alg:component}.

\subsection{Combination}
At every training iteration, RandMSAugment draws two operations randomly from [{\small \FFeatTransform, \FInterpolate, \FResizePaste}] . 

If the chosen operations are [{\small \FFeatTransform, \FInterpolate}],  we apply {\small \FFeatTransform} with the same randomly chosen parameters $\boldsymbol{\ell, \theta}$ to both images $\mathbf{x}_1$ and $\mathbf{x}_2$ and then {\small \FInterpolate} them. This is because interpolating images from different feature spaces can degrade objectness information, resulting in unrealistic images that are hard for the network to learn from.

For [{\small \FFeatTransform, \FResizePaste}] we
apply {\small \FFeatTransform} to $\mathbf{x}_1$ and $\mathbf{x}_2$
with independent randomly chosen parameters. Since patches from $\mathbf{x}_1$ and $\mathbf{x}_2$ are juxtaposed (and not interpolated), the information in them is preserved even when they come from different feature spaces. We add another layer of randomness by randomly choosing between the transformed image $\mathbf{x}^{ft}_{t}$ 
 and the original image $\mathbf{x}_t$. Images thus selected are combined with {\small \FResizePaste} with random $\lambda_{s}$.

\begin{figure}
\centering
\includegraphics[scale=0.25]{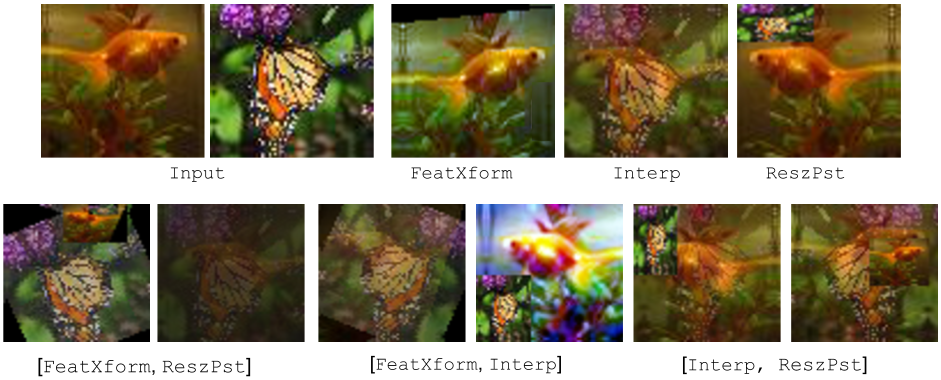} \\
\caption{The first row shows example input images and augmented samples from the component functions used in our method. The second row shows example augmentations for every pair of component operations, which are all samples from RandMSAugment. Our samples are diverse and exhibit real-world lighting and geometric changes.  }
\label{fig:mixed_samples}
\end{figure}

For [{\small \FInterpolate, \FResizePaste}], we first use {\small \FInterpolate} to make the interpolated image $\mathbf{x}^{ip}_{1,2}$. We then mix $\mathbf{x}^{ip}_{1,2}$ with one of  $\mathbf{x}_1$ or $\mathbf{x}_2$ using {\small \FResizePaste}. 
We would like $\mathbf{x}^{ip}_{1,2}$ to occupy a significant portion of the mixed sample. So, based on the value of the random scale factor $\lambda_{s}$, we make $\mathbf{x}^{ip}_{1,2}$ the first or second of the  samples to be mixed by {\small \FResizePaste}. 
Algorithm~\ref{alg:main} outlines these methods, with a visual description in Section~\ref{sec:visual_demo}. Figure~\ref{fig:mixed_samples} shows examples augmentations.

The two crucial aspects in RandMSAugment are the use of effective, complementary, yet simple component functions, and the extensive use of randomness in mixing. 
The precise details in combining the component functions are less important. We show in Section~\ref{sec:alternate} that an alternate way to combine [{\small \FInterpolate, \FResizePaste}]  has similar performance.

\SetKwProg{Proc}{Procedure}{}{}
\DontPrintSemicolon 

\begin{algorithm}
\smaller
\caption{RandMSAugment}
\label{alg:main}
\KwData{\(\mathbf{x}_1, \mathbf{x}_2 \gets \text{images} \), \(\mathbf{y}_1, \mathbf{y}_2 \gets \text{labels} \)}
\KwResult{\(\mathbf{x}^M_{1,2},\mathbf{y}^M_{1,2}\) }

\SetKwFunction{FRandMixAugment}{RandMixAugment}

\SetKwProg{Fn}{Function}{:}{}

\Fn{\FRandMixAugment{\(\mathbf{x}_1, \mathbf{x}_2,\mathbf{y}_1, \mathbf{y}_2\)}}{

\( \mathcal{O} \gets \) [\FFeatTransform, \FInterpolate, \FResizePaste] \;
\(\mathbf{o} \gets \) \FRandomChoice{\(\mathcal{O}, 2\)}\;
\If{\( \mathbf{o} \) in [\FFeatTransform, \FInterpolate] }{
\(\mathbf{x}^{ft}_1, \mathbf{y}^{ft}_1, \boldsymbol{\ell}, \boldsymbol{\theta} \gets \)
\FFeatTransform{\(\mathbf{x}_1, \mathbf{y}_1\)} \;
\(\mathbf{x}^{ft}_2, \mathbf{y}^{ft}_2
\gets \) \FFeatTransform{\(\mathbf{x}_2, \mathbf{y}_2, \boldsymbol{\ell}, \boldsymbol{\theta}\)} \;
\(\mathbf{x}^{M}_{1,2}, \mathbf{y}^{M}_{1,2} \gets \)
\FInterpolate{\(\mathbf{x}^{ft}_1, \mathbf{x}^{ft}_2,\mathbf{y}^{ft}_1, \mathbf{y}^{ft}_2\)}
    }

\If{\( \mathbf{o} \) in [\FFeatTransform, \FResizePaste] }{

 \For{$t = 1$ \KwTo $2$}{
\(\mathbf{x}^{ft}_t, \mathbf{y}^{ft}_t \gets \)
\FFeatTransform{\(\mathbf{x}_t, \mathbf{y}_t\)} \;
  $\mathbf{x}^{r}_t \gets $ \FRandomChoice{$[\mathbf{x}^{ft}_t, \mathbf{x}_t]$}\;
}
\(\mathbf{x}^{M}_{1,2}, \mathbf{y}^{M}_{1,2} \gets \)
\FResizePaste{\(\mathbf{x}^r_1, \mathbf{x}^r_2,\mathbf{y}^{ft}_1, \mathbf{y}^{ft}_2\)}
    }

\If{\( \mathbf{o} \) in [\FInterpolate, \FResizePaste] }{
\(\mathbf{x}^{ip}_{1,2}, \mathbf{y}^{ip}_{1,2} \gets \)
\FInterpolate{\(\mathbf{x}_1, \mathbf{x}_2,\mathbf{y}_1, \mathbf{y}_2\)} \;      ${\lambda_{s}} \gets $ \FRandomChoice{[0:1]}\;
\If{${\lambda_{s}} > 0.5$ }{
\(\mathbf{x}^{M}_{1,2}, \mathbf{y}^{M}_{1,2} \gets \) \FResizePaste{\(\mathbf{x}^{ip}_{1,2}, \mathbf{x}_2,\mathbf{y}^{ip}_{1,2}, \mathbf{y}_2, {\lambda_{s}}\)}
}
\Else{\(\mathbf{x}^{M}_{1,2}, \mathbf{y}^{M}_{1,2} \gets \) \FResizePaste{\(\mathbf{x}_1, \mathbf{x}^{ip}_{1,2}, \mathbf{y}_1,\mathbf{y}^{ip}_{1,2}, {\lambda_{s}}\)}
}
    \Return{\(\mathbf{x}^{M}_{1,2}, \mathbf{y}^{M}_{1,2}\)}\;
}
}
\end{algorithm}

\section{Experiments}
\label{sec:expt}
\SetKwFunction{FRP}{RP}
\SetKwFunction{FIP}{IP}
\SetKwFunction{FFT}{FT}
\definecolor{customgreen}{rgb}{0, 0.4, 0.}

We evaluate RandMSAugment for image classification on STL-10~\cite{coates2011analysis}, CIFAR-100~\cite{krizhevsky2009learning}, and Tiny-Imagenet datasets~\cite{chrabaszcz2017downsampled}, with 5,000, 50,000, and 100,000 training samples respectively.
We run experiments on the full and limited data settings. For limited data, we randomly sample various amounts from the full dataset in a class-balanced manner, similar to SSL methods\cite{sohn2020fixmatch, zhang2021flexmatch}. However, unlike SSL methods, we do not use unlabelled data, as our goal is to understand and leverage the potential of \emph{fully supervised} methods for \emph{sparse data} problems. Using  unlabelled data is complementary to our approach.  

\paragraph{Methods}
We compare RandMSAugment to recent SotA techniques that use no additional data for augmentation:  Basic (flips and crops), Mixup~\cite{zhang2017mixup},  Cutmix~\cite{yun2019cutmix}, Preset-RandAugment (Section~\ref{sec:intro}), ResizeMix~\cite{qin2020resizemix}, Augmix \cite{hendrycks2019augmix}, AutoMix~\cite{zhu2020automix}, HMix and GMix~\cite{park2022unified}. The neural module in Automix requires several thousands more parameters than the baseline  architecture used in other methods (Table~\ref{sec:automix}), as indicated with $^*$ in the comparison tables.


\paragraph{Settings}
For a fair comparison, we ran experiments for all methods with the same training settings~\cite{oliver2018realistic} in TorchSSL~\cite{zhang2021flexmatch}. Following standard practice~\cite{cubuk2020randaugment},  
we use a WRN 28-10 \cite{zagoruyko2016wide} for CIFAR-100, WRN 28-2 variant \cite{pmlr-v119-zhou20d} for STL-10 and PreAct-ResNet18 \cite{he2016identity} for Tiny-Imagenet. 
We train models for $2^{20}$ iterations when using all the data. With sparse data, we train  for 500K iterations, to be fair to methods with different convergence speeds. For each method we report the best accuracy from the last 20 snapshots~\cite{zhang2021flexmatch}. 
 We use SGD with momentum 0.9, weight decay 0.0005, and batch size 64. We use the MultistepLR schedule with an initial learning rate 0.03, gamma 0.3 and step ratio 0.1. We perform an exponential moving average with decay factor 0.999. 

 For the component augmentations used in our method, we use default parameters specified by the methods that proposed them. We use the same set of low-level transforms specified in Preset-RandAugment, with N=2 and randomly sample M from the full range of magnitudes. For Mixup, we draw $\lambda$ from a Beta distribution with $\alpha=1.0$ for CIFAR-100 and STL-10, and $\alpha=0.2$ for Tiny-Imagenet, as in \cite{zhang2017mixup}. The ResizeMix resize-factor is random in (0.2,0.8).

\subsection{Performance}
We compare the methods in terms of the Top-1 accuracy rate, and report the mean and standard deviation over 3 independent networks trained with different random seeds in Tables~\ref{tab:cifar}-\ref{tab:tiny}. The best method is shown in bold and the second best is underlined. We also show results of models trained with our intermediate combinations:  [{\small \FFeatTransform, \FResizePaste}] abbreviated as [\FFT, \FRP], [{\small \FFeatTransform, \FInterpolate}]  abbreviated as [\FFT, \FIP],
and RandMSAugment$^{-}$, which randomly chooses between [\FFT, \FRP] and [\FFT, \FIP]. This combination emphasizes  {\small \FFeatTransform}, which we showed in Section~\ref{sec:properties} to be effective on sparse training data. 

Our RandMSAugment performs the best on all 3 datasets. We also show the gain in performance in terms of percentage points over the second best method (underlined). The gains are most dramatic for sparse samples (4, 25, 100 samples / class), ranging from 4.1\% (Tiny-Imagenet) to 6.75\% (STL-10). We also improve performance when the samples are abundant (500 samples / class), with gains ranging from 1.03 \% (CIFAR-100) to 2.47 \% (STL-10). 

STL-10 has only 10 classes, so there are only 40 samples in total for the sparsest case (versus 400 in CIFAR-100 and 800 in Tiny-Imagenet). Here, the {\small \FFeatTransform} based method, Preset-RandAugment performs especially well, and so emphasizing {\small \FFeatTransform} in RandMSAugment$^{-}$ is more beneficial than  RandMSAugment. 

\begin{table}
    \centering
    \scriptsize
    \caption{Top-1 accuracy (\%, $\uparrow$) on CIFAR-100 (100 classes). [\protect\FFT, \protect\FRP] and [\protect\FFT, \protect\FIP] are our intermediate outputs.  }
    \label{tab:cifar}
    \begin{tabular}{lcccc} 
        \toprule
      Augmentation  & \multicolumn{4}{c}{Samples / Class}  \\ 
        & 4 & 25 & 100 & 500  \\
        \midrule
Basic & 9.93\tiny{$\pm$1.36} & 41.57\tiny{$\pm$0.08} & 63.30\tiny{$\pm$0.10} & 81.16\tiny{$\pm$0.03}\\

Cutmix & 14.50\tiny{$\pm$0.04} & 44.27\tiny{$\pm$0.28} & 69.80\tiny{$\pm$0.06} & 84.95\tiny{$\pm$0.15}\\

ResizeMix & 13.84\tiny{$\pm$0.18} & 45.74\tiny{$\pm$0.47} & \underline{70.09\tiny{$\pm$0.23}} & \underline{85.25\tiny{$\pm$0.19}}\\

Mixup & 17.31\tiny{$\pm$0.17} & 47.87\tiny{$\pm$0.10} & 67.21\tiny{$\pm$0.04} & 82.93\tiny{$\pm$0.20}\\

{\smaller Preset-RandAugment} & 19.74\tiny{$\pm$0.25} & 48.21\tiny{$\pm$0.18} & 67.69\tiny{$\pm$0.04} & 83.43\tiny{$\pm$0.10}\\

Augmix & 14.58\tiny{$\pm$0.44} & 40.53\tiny{$\pm$0.55} & 64.02\tiny{$\pm$0.16} & 81.29\tiny{$\pm$0.06}\\

HMix & 18.57\tiny{$\pm$0.13} & \underline{50.06\tiny{$\pm$0.05}} & 69.20\tiny{$\pm$0.09} & 84.31\tiny{$\pm$0.10}\\

GMix & \underline{20.02\tiny{$\pm$0.11}} & 48.34\tiny{$\pm$0.04} & 68.89\tiny{$\pm$0.24} & 84.24\tiny{$\pm$0.12}\\

AutoMix$^*$ & 16.16\tiny{$\pm$0.01} & 48.02\tiny{$\pm$1.29} & 69.03\tiny{$\pm$0.02} & 85.03\tiny{$\pm$0.24}\\
\hline

 [\protect\FFT, \protect\FRP] & 22.21\tiny{$\pm$0.05} & 53.29\tiny{$\pm$0.17} & 72.19\tiny{$\pm$0.08} & 86.13\tiny{$\pm$0.13}\\

 [\protect\FFT, \protect\FIP] & 21.58\tiny{$\pm$0.19} & 51.45\tiny{$\pm$0.07} & 69.64\tiny{$\pm$0.15} & 84.14\tiny{$\pm$0.29}\\

 RandMSAugment$^{-}$ & \bf{25.07\tiny{$\pm$0.10}} & 53.87\tiny{$\pm$0.02} & 72.04\tiny{$\pm$0.07} & 85.87\tiny{$\pm$0.03}\\

 RandMSAugment & 24.27\tiny{$\pm$0.26} & \bf{54.44\tiny{$\pm$0.06}} & \bf{72.67\tiny{$\pm$0.03}} & \bf{86.28\tiny{$\pm$0.24}}\\
\bottomrule

\textcolor{customgreen}{Gains} & \textcolor{customgreen}{\bf + 5.06\tiny{$\pm$0.21}} & \textcolor{customgreen}{\bf + 4.37\tiny{$\pm$0.02}} & \textcolor{customgreen}{\bf + 2.58\tiny{$\pm$0.25}} & \textcolor{customgreen}{\bf + 1.03\tiny{$\pm$0.29}}\\
\bottomrule
\end{tabular}
\end{table}

\begin{table}
\scriptsize
    \centering
    \caption{Top-1 accuracy (\%, $\uparrow$) on STL-10 (10 classes).}
    \label{tab:stl}    
    \begin{tabular}{lcccc} 
        \toprule
      Augmentation  & \multicolumn{4}{c}{Samples / Class}  \\ 
        & 4 & 25 & 100 & 500  \\  
        \midrule
Basic & 22.28\tiny{$\pm$0.70} & 37.03\tiny{$\pm$0.56} & 66.37\tiny{$\pm$0.26} & 85.77\tiny{$\pm$0.20}\\

Cutmix & 24.17\tiny{$\pm$0.52} & 47.88\tiny{$\pm$0.34} & 68.01\tiny{$\pm$0.54} & 90.62\tiny{$\pm$0.12}\\

ResizeMix & 23.59\tiny{$\pm$0.49} & 47.78\tiny{$\pm$0.35} & 70.23\tiny{$\pm$0.26} & 91.25\tiny{$\pm$0.07}\\

Mixup & 29.40\tiny{$\pm$0.16} & 55.35\tiny{$\pm$0.35} & 73.06\tiny{$\pm$0.15} & 89.88\tiny{$\pm$0.07}\\

{\smaller Preset-RandAugment} & \underline{39.77\tiny{$\pm$0.17}} & \underline{63.40\tiny{$\pm$0.43}} & \underline{77.88\tiny{$\pm$0.14}} & 91.21\tiny{$\pm$0.13}\\

Augmix & 29.35\tiny{$\pm$0.43} & 50.10\tiny{$\pm$0.37} & 66.50\tiny{$\pm$0.09} & 87.05\tiny{$\pm$0.10}\\

HMix & 30.43\tiny{$\pm$0.39} & 57.26\tiny{$\pm$0.06} & 75.99\tiny{$\pm$0.19} & \underline{91.58\tiny{$\pm$0.09}}\\

GMix & 31.66\tiny{$\pm$0.19} & 57.39\tiny{$\pm$0.23} & 74.04\tiny{$\pm$0.46} & 90.82\tiny{$\pm$0.03}\\
AutoMix$^*$ & 33.39\tiny{$\pm$0.19} & 56.05\tiny{$\pm$0.26} & 74.21\tiny{$\pm$0.14} & 90.74\tiny{$\pm$0.05}\\ 
\hline
[\protect\FFT, \protect\FRP]  & 42.56\tiny{$\pm$0.70} & 67.97\tiny{$\pm$0.42} & 82.70\tiny{$\pm$0.15} & 93.24\tiny{$\pm$0.16}\\

[\protect\FFT, \protect\FIP]  & 43.70\tiny{$\pm$0.37} & 64.32\tiny{$\pm$0.24} & 80.47\tiny{$\pm$0.21} & 92.55\tiny{$\pm$0.17}\\

RandMSAugment$^{-}$ & \bf{45.01\tiny{$\pm$0.28}} & \bf{70.14\tiny{$\pm$0.51}} & \bf{83.81\tiny{$\pm$0.12}} & 94.02\tiny{$\pm$0.06}\\

RandMSAugment & 43.50\tiny{$\pm$0.35} & 68.83\tiny{$\pm$0.07} & 83.67\tiny{$\pm$0.17} & \bf{94.05\tiny{$\pm$0.05}}\\
\bottomrule
\textcolor{customgreen}{Gains} & \textcolor{customgreen}{\bf{+ 5.24\tiny{$\pm$0.14}}}& \textcolor{customgreen}{\bf{+ 6.75\tiny{$\pm$0.84}}}& \textcolor{customgreen}{\bf{+ 5.94\tiny{$\pm$0.26}}}& \textcolor{customgreen}{\bf{+ 2.47\tiny{$\pm$0.06}}}\\
    \bottomrule
    \end{tabular}
\end{table}

\begin{table}
\scriptsize
    \centering
    \caption{Top-1 accuracy (\%, $\uparrow$) on Tiny-Imagenet (200 classes). }
      \label{tab:tiny}   
    \begin{tabular}{lcccc} 
        \toprule
      Augmentation  & \multicolumn{4}{c}{Samples / Class}  \\ 
       & 4 & 25 & 100 & 500  \\
        \midrule
Basic & 6.08\tiny{$\pm$0.23} & 25.79\tiny{$\pm$0.14} & 45.03\tiny{$\pm$0.13} & 64.40\tiny{$\pm$0.17}\\

Cutmix & 8.13\tiny{$\pm$0.08} & 25.85\tiny{$\pm$0.05} & 45.20\tiny{$\pm$0.83} & 67.52\tiny{$\pm$0.04}\\

ResizeMix & 7.80\tiny{$\pm$0.10} & 25.69\tiny{$\pm$0.13} & 46.06\tiny{$\pm$0.57} & \underline{67.77\tiny{$\pm$0.22}}\\

Mixup & 7.92\tiny{$\pm$0.16} & 27.30\tiny{$\pm$0.46} & 47.91\tiny{$\pm$0.24} & 66.02\tiny{$\pm$0.25}\\

{\smaller Preset-RandAugment} & \underline{10.54\tiny{$\pm$0.13}} & \underline{29.43\tiny{$\pm$0.10}} & \underline{49.00\tiny{$\pm$0.15}} & 67.47\tiny{$\pm$0.10}\\

Augmix & 9.51\tiny{$\pm$0.10} & 26.51\tiny{$\pm$0.26} & 45.29\tiny{$\pm$0.47} & 64.57\tiny{$\pm$0.28}\\

HMix & 7.96\tiny{$\pm$0.31} & 26.80\tiny{$\pm$0.89} & 46.11\tiny{$\pm$0.34} & 67.71\tiny{$\pm$0.42}\\

GMix & 10.05\tiny{$\pm$0.17} & 26.21\tiny{$\pm$0.20} & 46.49\tiny{$\pm$0.40} & 67.61\tiny{$\pm$0.34}\\

Automix$^*$ & 8.14\tiny{$\pm$0.36} & 23.30\tiny{$\pm$0.29} & 45.75\tiny{$\pm$0.50} & 66.81\tiny{$\pm$0.20}\\
\hline 
[\protect\FFT, \protect\FRP] & 11.62\tiny{$\pm$0.15} & 31.94\tiny{$\pm$0.15} & 49.70\tiny{$\pm$0.25} & 69.36\tiny{$\pm$0.33}\\

[\protect\FFT, \protect\FIP] & 11.53\tiny{$\pm$0.10} & 32.03\tiny{$\pm$0.41} & 50.82\tiny{$\pm$0.10} & 67.81\tiny{$\pm$0.29}\\

RandMSAugment$^-$ & \bf{14.63\tiny{$\pm$0.31}} & 33.88\tiny{$\pm$0.13} & 52.11\tiny{$\pm$0.50} & 69.70\tiny{$\pm$0.05}\\

RandMSAugment & 14.48\tiny{$\pm$0.27} & \bf{34.61\tiny{$\pm$0.22}} & \bf{52.37\tiny{$\pm$0.09}} & \bf{69.96\tiny{$\pm$0.27}}\\
\bottomrule
Gains& \textcolor{customgreen}{\bf{+ 4.10\tiny{$\pm$0.39}}}& \textcolor{customgreen}{\bf{+ 5.18\tiny{$\pm$0.30}}}& \textcolor{customgreen}{\bf{+ 3.37\tiny{$\pm$0.23}}}& \textcolor{customgreen}{\bf{+ 2.19\tiny{$\pm$0.13}}}\\

\bottomrule
    \end{tabular}
\end{table}

\subsection{Ablation study}
We compare the relative importance of the different components of our method via Top-1 accuracy in Table~\ref{tab:ablation}. The top part of the table shows that that {\small \FFeatTransform} is the most effective individual component. Among pairwise combinations (middle part), [{\small \FFeatTransform, \FResizePaste}] works best. The bottom part shows that balancing the 3 components yields the best results versus emphasizing any one of them.

We illustrate the augmentation properties (Section~\ref{sec:properties}) of RandMSAugment, and show additional experiments with different parameters and training settings in Sections~\ref{sec:randms_properties}, \ref{sec:param} and \ref{sec:training} respectively.
Code is also attached with the Supplementary material. 

\begin{table}
\scriptsize
    \centering
    \caption{Top-1 accuracy (\%, $\uparrow$) on CIFAR-100 for 25 samples / class. Orange ticks flag emphasized components in the mix.}
      \label{tab:ablation}   
    \begin{tabular}{cccc} 
        \toprule
     \smaller \FFeatTransform & \smaller \FResizePaste & \smaller \FInterpolate & Top-1 Accuracy \\
        \midrule

 \checkmark & & & 48.32\tiny{$\pm$0.36} \\ 
  & \checkmark && 46.62\tiny{$\pm$0.30} \\ 
 & & \checkmark & 47.85\tiny{$\pm$0.19}\\ 
 \hline
  \checkmark & \checkmark &&  53.29\tiny{$\pm$0.17} \\ 
  \checkmark & & \checkmark & 51.45\tiny{$\pm$0.07} \\ 
  \checkmark & \checkmark && 50.23\tiny{$\pm$0.09} \\ 
\hline 
\textcolor{BurntOrange}{\checkmark} & \checkmark & \checkmark & 53.83\tiny{$\pm$0.03} \\
\checkmark  & \textcolor{BurntOrange}{\checkmark} & \checkmark & 54.39\tiny{$\pm$0.08} \\ 
\checkmark  & \checkmark  & \textcolor{BurntOrange}{\checkmark} & 53.30 \tiny{$\pm$0.11} \\ 
{\checkmark}  & {\checkmark}  & {\checkmark} & \bf{54.46\tiny{$\pm$0.05}} \\

\bottomrule
    \end{tabular}
\end{table}

\section{Conclusion}
We evaluated foundational augmentation methods for data-scarce scenarios based on 3 important properties: diversity, realism, and the one we introduced: fast convergence via feature invariance. We showed that Preset-RandAugment strikes an optimal balance among them, using low-level feature transforms as its key component. 
 We proposed a novel technique, RandMSAugment, which combines these low-level transforms with the complementary strengths of  MSDA methods. We showed through extensive comparisons over 3 datasets and ablation studies that our efficient method produces dramatic gains over the competition for sparse samples and more moderate gains for abundant data. 
Our method operates in the fully-supervised scenario. 
We plan to explore RandMSAugment for SSL in future work.

{
    \small
    \bibliographystyle{ieeenat_fullname}
    \bibliography{main}
}

\clearpage
\setcounter{page}{1}
\maketitlesupplementary
\newpage

\section{Additional Experiments}

\subsection{Visual Description of Algorithm}
\label{sec:visual_demo}
\begin{figure}[th]
\centering
\includegraphics[scale=0.13]{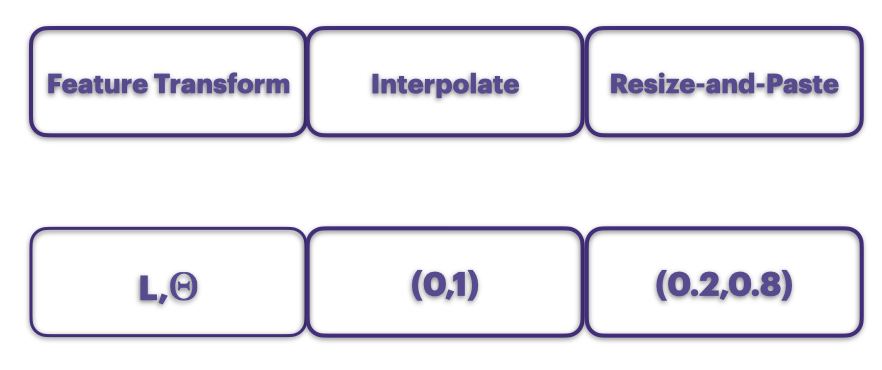} \\
(a) \\
 \renewcommand{\arraystretch}{0.7}
\includegraphics[scale=0.1]{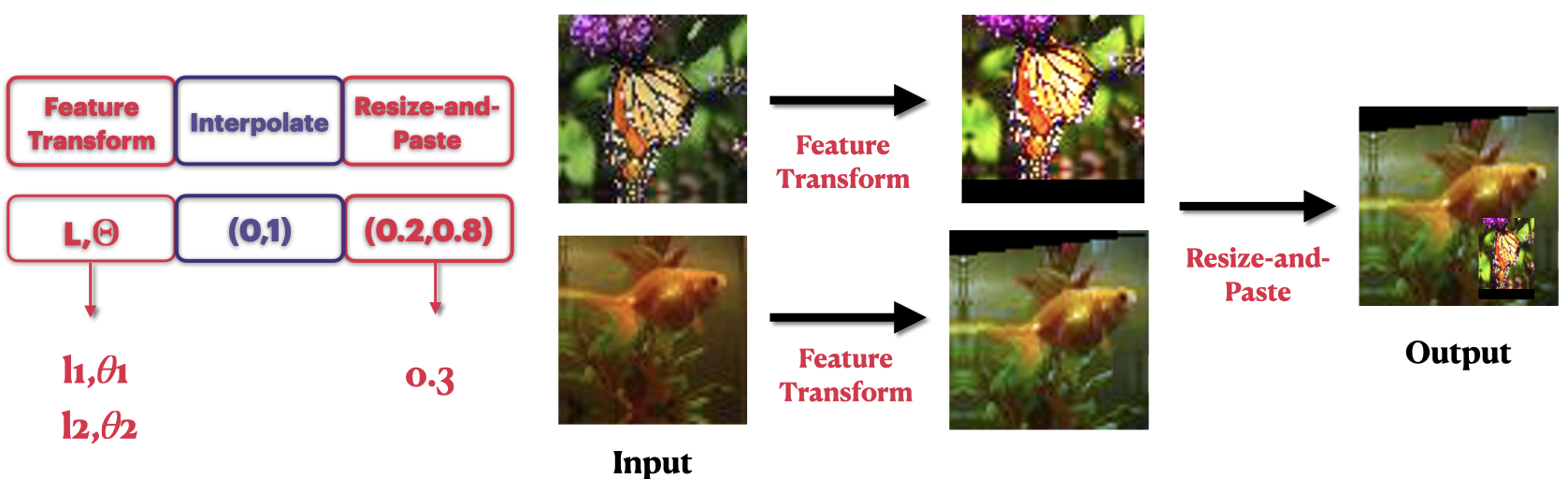} \\
(b) \\
\includegraphics[scale=0.1]{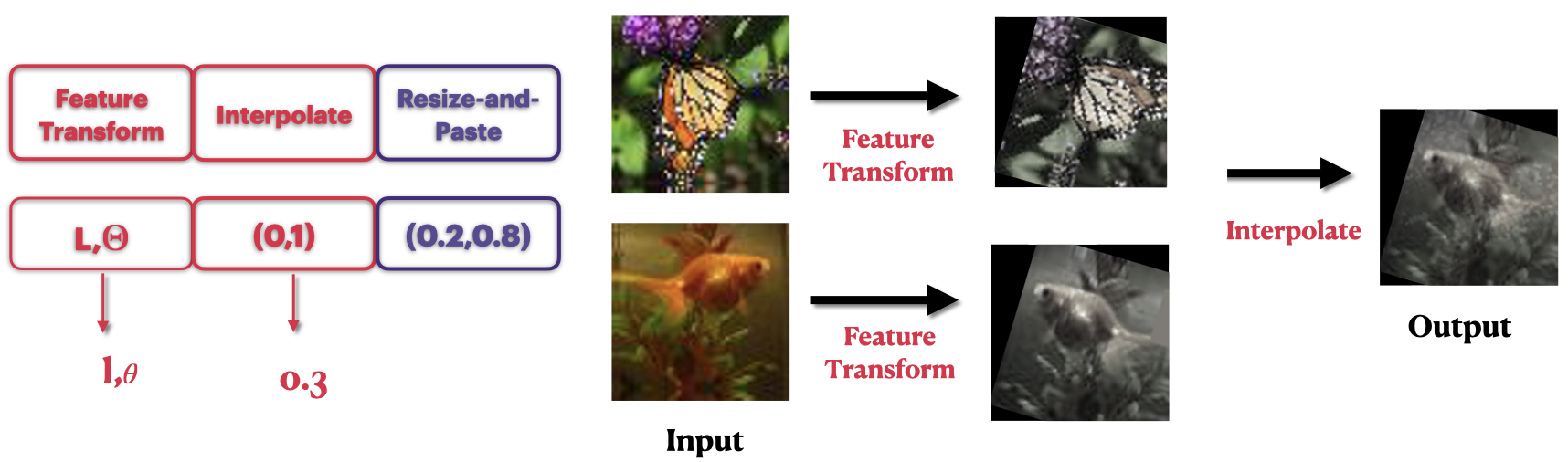} \\
(c) \\
\includegraphics[scale=0.1]{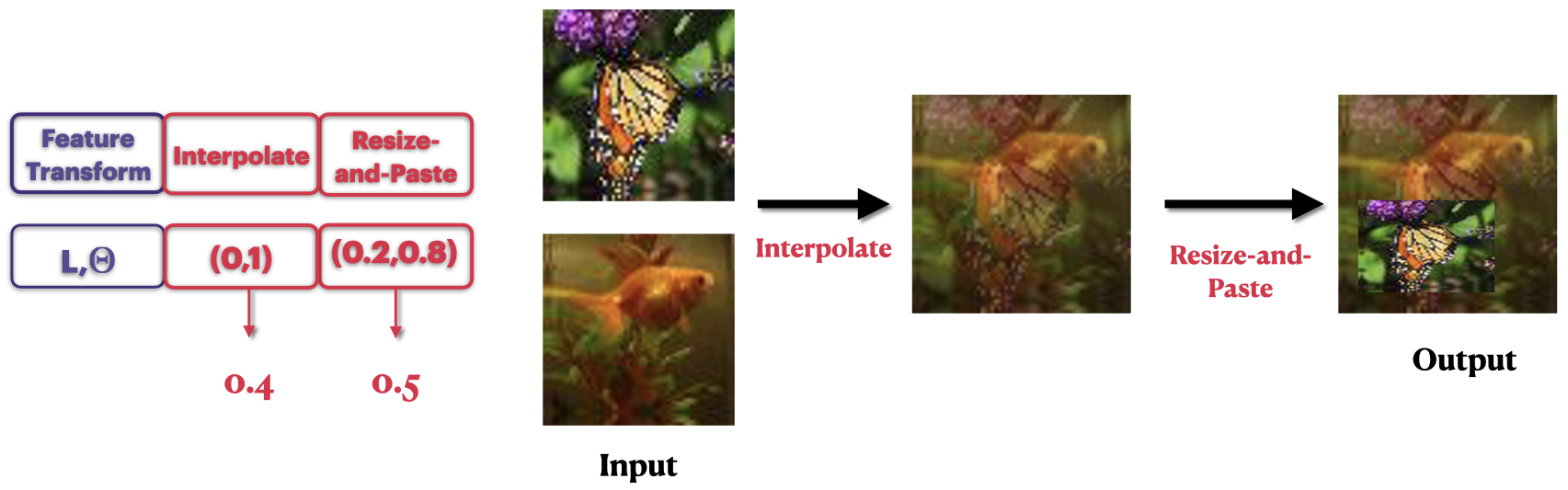} \\
(d) \\
\caption{(a) The different component operations in RandMSAugment. In each training iteration, we  randomly choose two operations out of them. Their associated parameters are also chosen from  pre-defined ranges,  uniformly at random.  We show an example for [\protect\FFeatTransform, \protect\FResizePaste] in (b), [\protect\FFeatTransform, \protect\FInterpolate] in (c), and [\protect\FInterpolate, \protect\FResizePaste] in (d). } 
\label{fig:vis_algo}
\end{figure}

Figure~\ref{fig:vis_algo} is a visual demonstration of Algorithm~\ref{alg:main}.

\subsection{Alternate Way to Combine [{\small \protect\FInterpolate, \protect\FResizePaste}]}
\label{sec:alternate}
The two crucial aspects in RandMSAugment are the use of effective, complementary, yet simple component functions, and the extensive use of randomness in mixing. 
The precise details in combining the component functions are less important. We demonstrate this by considering an alternate way to combine [{\small \FInterpolate, \FResizePaste}], where we first compute the interpolated image $\mathbf{x}^{ip}_{1,2}$. Then, we add stochasticity to our procedure by randomly deciding  between $\mathbf{x}^{ip}_{1,2}$  and the original image $\mathbf{x}_t$ for $t={1,2}$. The images selected from this step are combined using {\small \FResizePaste} with a random $\lambda_{s}$. The performance of both the main [{\small \FInterpolate, \FResizePaste}] function described in Algorithm~\ref{alg:main} and the alternate function are compared in Table~\ref{tab:alternate} using the Tiny-Imagenet dataset, where the details of the training settings is in Section~\ref{sec:expt}. Both the main and the alternate functions yield similar performance, illustrating that the exact mechanism of combining the component functions is not crucial.

\begin{table}[h]
    \centering
    \scriptsize
  \begin{tabular}{lcccc} 
        \toprule
      Augmentation  & \multicolumn{4}{c}{Samples / Class}  \\ 
        & 4 & 25 & 100 & 500  \\
        \midrule
      Main (Alg~\ref{alg:main}) &  11.81\tiny{$\pm$0.12} & 32.16\tiny{$\pm$0.21} & 49.59\tiny{$\pm$0.49} & 69.50\tiny{$\pm$0.19} \\ 
       Alternate & 12.01\tiny{$\pm$0.25} & 31.97\tiny{$\pm$0.47} & 50.19\tiny{$\pm$0.18} & 69.28\tiny{$\pm$0.21} \\ 
        
  \bottomrule
        \end{tabular}
            \caption{Two different ways of combining [\protect\FInterpolate, \protect\FResizePaste] yield similar performance for different number of samples per class for the Tiny-Imagenet dataset with 200 classes.  }

    \label{tab:alternate}
  \end{table}
  
\subsection{AutoMix}
\label{sec:automix}
Automix uses hundreds of thousands of extra parameters compared to the baseline  network architecture used by all the other methods, as shown in Table~\ref{tab:automix}. Comparing Automix with other methods trained on models with fewer parameters is unfair, but we show this comparison because other published works do.

\begin{table}[h]
\scriptsize
    \centering
    \begin{tabular}{|c|ccc|}
    \toprule
        Network & WRN 28-10 & Preact-Resnet18 & WRN-Var 28-2\\
        \midrule
        Baseline &  36.55 M & 11.27 M & 59.34\\
        Automix &  36.66 M & 11.40 M & 59.91\\
         \midrule
         Extra & 115,781 & 135,753 & 57,215 \\
         \bottomrule         
    \end{tabular}
    \caption{Automix uses about 57-136K more parameters compared to the baseline models used in CIFAR100, Tiny-Imagenet and STL-10 respectively.   }
    \label{tab:automix}
\end{table}

\subsection{Properties of RandMSAugment}
\label{sec:randms_properties}
\begin{figure*}[h]
\centering
\begin{tabular}{@{}c@{}c@{}}
\includegraphics[scale=0.27]{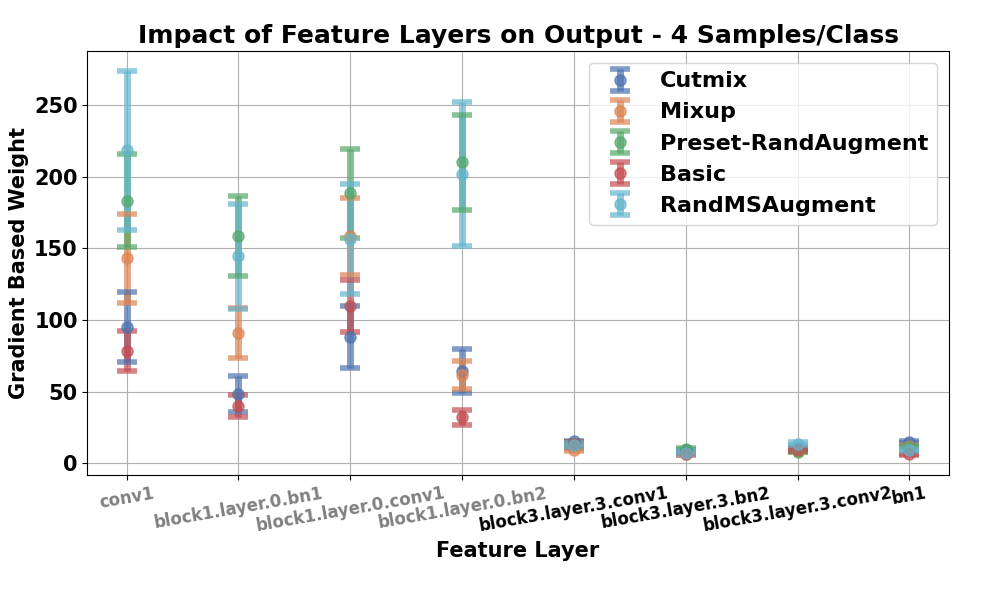} &
\includegraphics[scale=0.28]{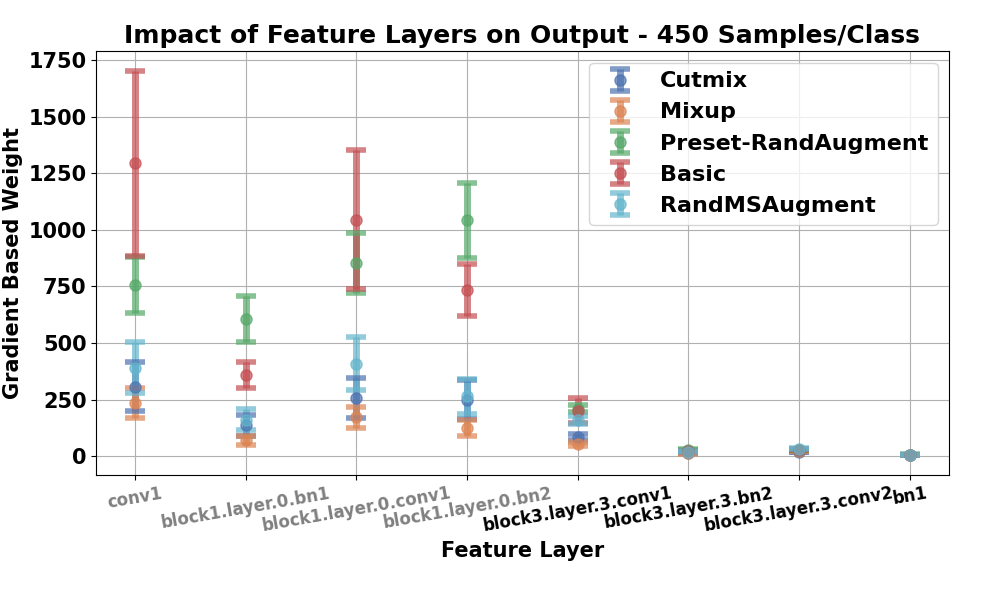} \\
(a) & (b) \\
\includegraphics[scale=0.27]{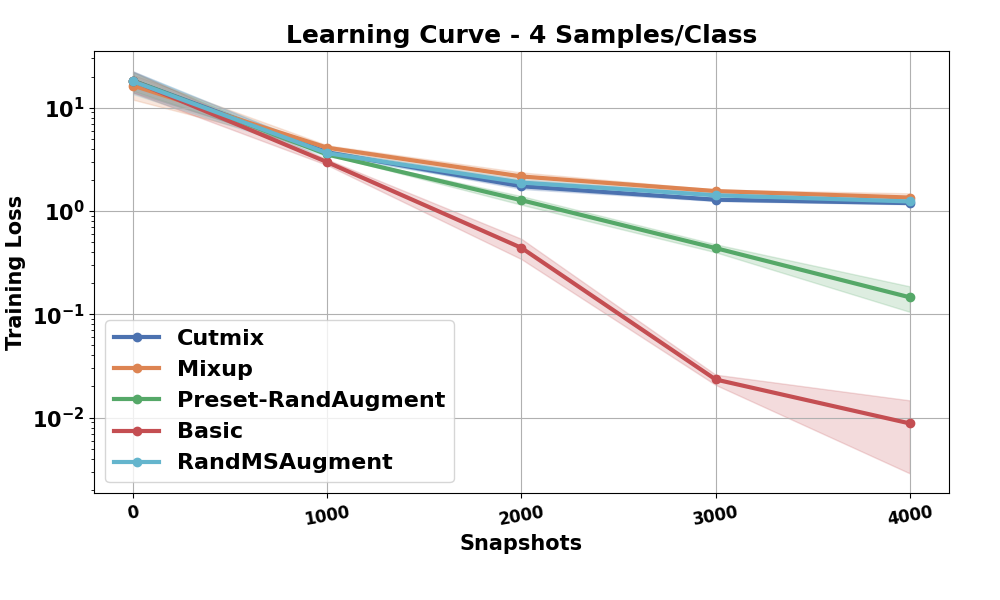} &
\includegraphics[scale=0.27]{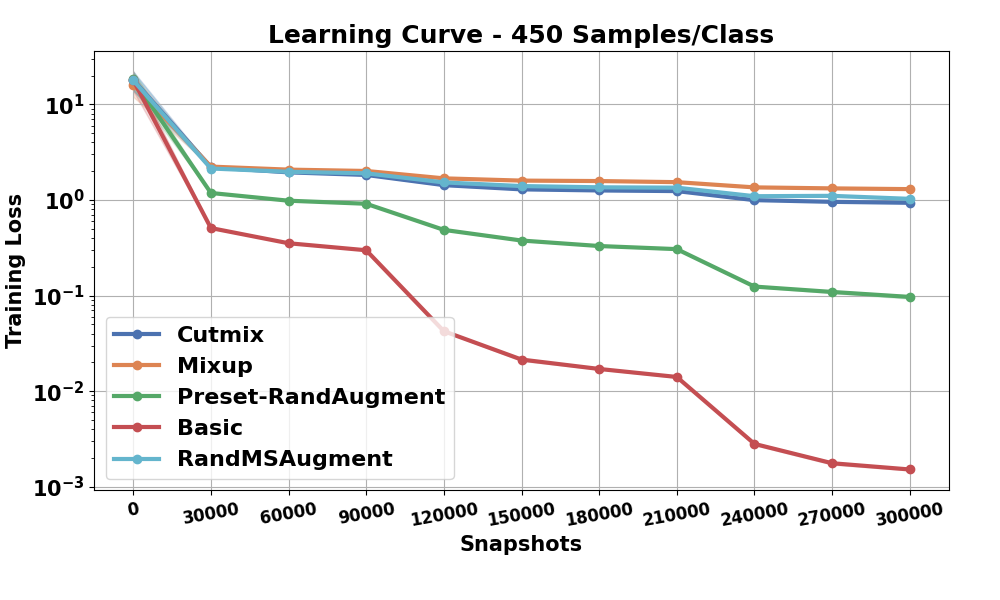}  \\
(c) & (d) \\
\end{tabular}

\caption{(a-b) Gradient based weights $w_r^c$ averaged over all  samples  $(\mathbf{x},\mathbf{y})  \in \mathcal{V}_e$, for 4  lowest (gray) and highest (black) feature maps in the model. The weights for RandMSAugment is especially high for sparse labels. (c-d) Learning curve for different augmentation methods is shown, where RandMSAugment's convergence speed is closer to MSDA methods.
Averages are computed over 3 models trained with different random seeds.}
\label{fig:lc_randms}
\end{figure*}

\begin{figure}[h]
\centering
\includegraphics[scale=0.25]{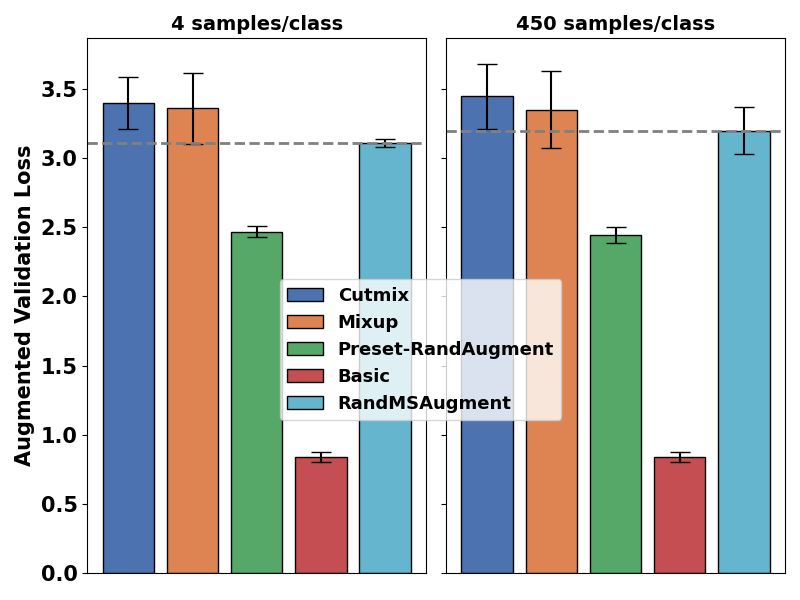} \\
\caption{Distributional similarity of RandMSAugment samples to $P_e$, compared to  samples from different augmentations. The realism is RandMSAugment samples is between Preset-RandAugment and MSDA methods. Error bars are computed over 3 models trained with different random seeds. }
\label{fig:similarity_randms}
\end{figure}

We compare RandMSAugment with foundational augmentation methods in terms of the properties defined in Section~\ref{sec:properties}. RandMSAugment's  diversity is comparable to Preset-RandAugment as seen in Table~\ref{tab:diversity_randms}. Further, its use of low-level transforms enables it to learn good low-level features (Figure~\ref{fig:lc_randms}(a-b)), which are especially impactful for sparse samples. The realism in RandMSAugment samples is between Preset-RandAugment and MSDA  samples (Figure~\ref{fig:similarity_randms} and ~\ref{fig:mixed_samples}), and so is its convergence rate (Figure~\ref{fig:lc_randms}(c-d)). 

Overall, RandMSAugment samples are closer to the feature transform based method, Preset-RandAugment, in terms of diversity and encouraging the learning of invariant low-level features, and closer to MSDA methods with respect to realism and speed of convergence. 
In other words, RandMSAugment's mixed samples sacrifice some realism in return for good regularization effects. 

\begin{table}[h]
\scriptsize
    \centering
    \begin{tabular}{|c|c|}
          \hline
            Method & Diversity \\
            \hline \hline
            RandMSAugment & 2386.44 \\
            Preset-RandAugment & {\bf{2516.78}}\\
           Cutmix & 1775.41\\
           Mixup & 1643.86\\
           Basic & 1141.87\\
             \hline
        \end{tabular}
  \caption{Diversity $ \langle{\sigma_a^{2}} \rangle$ of RandMSAugment compared to  foundational augmentation methods on the CIFAR100 dataset. RandMSAugment's diversity is comparable to Preset-RandAugment.}
    \label{tab:diversity_randms}
  \end{table}

\subsection{Ablation Study with Parameter $\alpha$}
\label{sec:param}
Next, we empirically analyse the parameter $\alpha$ that is used to modify the shape of the Beta distribution, from which the interpolation parameter $\lambda$ is drawn.
When $\alpha<1$, the distribution is skewed towards 0, whereas $\alpha=1$ produces a uniform distribution.
We compare the performance of Mixup and RandMSAugment for two values of the parameter $\alpha$
recommended in \citep{zhang2017mixup} in Table~\ref{tab:alpha}. The differences in the performance between the two cases does not seem to be statistically significant. 

\begin{table*}[h]
\scriptsize
    \centering
    \begin{tabular}{lcccc} 
        \toprule
      Augmentation  & \multicolumn{4}{c}{Samples / Class}  \\ 
       & 4 & 25 & 100 & 500  \\
        \midrule
Mixup $\alpha=0.2$ & 5.39\tiny{$\pm$3.52 } &  \bf{27.30\tiny{$\pm$0.46 }} &  \bf{47.91\tiny{$\pm$0.24 }} &  \bf{66.02\tiny{$\pm$0.25 }} \\
Mixup $\alpha=1.0$ & \bf{7.79\tiny{$\pm$0.19 }} & 25.18\tiny{$\pm$0.30 } & 45.85\tiny{$\pm$0.69 } & 65.89\tiny{$\pm$0.17 }  \\
\hline
 RandMSAugment $\alpha=0.2$ & \bf{14.48\tiny{$\pm$0.27 }} &  \bf{34.61\tiny{$\pm$0.23 }} &  \bf{52.37\tiny{$\pm$0.08 }} &  69.96\tiny{$\pm$0.27 }  \\
 RandMSAugment $\alpha=1.0$ & 13.12\tiny{$\pm$0.47 } &  33.90\tiny{$\pm$0.35 } &  52.22\tiny{$\pm$0.16 } &  \bf{70.64\tiny{$\pm$0.16 }} \\

\bottomrule
    \end{tabular}
    \caption{Comparison between results obtained with  $\alpha=0.2$ and $\alpha=1.0$  for Mixup and RandMSAugment in terms of Top-1 accuracy (\%, $\uparrow$) on Tiny-Imagenet. In most cases, the standard error across corresponding results overlaps and is therefore not statistically significant. We use $\alpha=0.2$ in our main experiments as recommended by the Mixup paper~\cite{zhang2017mixup} for Tiny-Imagenet, which seems to work better overall with both Mixup and RandMSAugment. }
    \label{tab:alpha}   
\end{table*}

\subsection{Ablation Study with Different Learning Rates}
\label{sec:training}

Finally, we show that the difference between results obtained with slightly different training settings is small. Some methods~\citep{kim2020puzzle} use a learning rate of 0.1 to train Tiny-Imagenet. 
Table~\ref{tab:lr} shows that the difference in performance between using a learning rate of 0.1 and 0.03 on  Tiny-Imagenet is under 1\% in most cases, and that a learning rate of 0.03 does best overall. 

\begin{table*}[th]
\scriptsize
    \centering
    \begin{tabular}{lcccc} 
        \toprule
      Augmentation  & \multicolumn{4}{c}{Samples / Class}  \\ 
       & 4 & 25 & 100 & 500  \\
        \midrule
{\smaller Preset-RandAugment} $lr=0.1$  & \bf{11.63\tiny{$\pm$0.16 }} &  \bf{32.62\tiny{$\pm$0.03 }} &  {49.05\tiny{$\pm$0.05 }} &  66.59\tiny{$\pm$0.30 }  \\

{\smaller Preset-RandAugment} $lr=0.03$ & {10.54\tiny{$\pm$0.13}} & {29.43\tiny{$\pm$0.10}} & \bf{49.20\tiny{$\pm$0.15}} & \bf{67.47\tiny{$\pm$0.10}}\\

\hline

ResizeMix $lr=0.1$ & \bf{8.53\tiny{$\pm$0.05 }} &  24.16\tiny{$\pm$0.12 } &  45.97\tiny{$\pm$0.04 } &  66.56\tiny{$\pm$0.43 }  \\

ResizeMix $lr=0.03$ & 7.80\tiny{$\pm$0.10} & \bf{25.69\tiny{$\pm$0.13}} & \bf{46.06\tiny{$\pm$0.57}} & \bf{67.77\tiny{$\pm$0.22}}\\
\hline 

Mixup $lr=0.1$ & 7.79\tiny{$\pm$0.11 }  &  25.38\tiny{$\pm$0.64 } &  44.30\tiny{$\pm$0.03 } &  65.52\tiny{$\pm$0.08 } \\

Mixup $lr=0.03$ & \bf{7.92\tiny{$\pm$0.16}} & \bf{27.30\tiny{$\pm$0.46}} & \bf{47.91\tiny{$\pm$0.24}} & \bf{66.02\tiny{$\pm$0.25}}\\

\hline
RandMSAugment $lr=0.1$ & {14.21\tiny{$\pm$0.00 }} &  {34.30\tiny{$\pm$0.34 }} &  \bf{52.42\tiny{$\pm$0.41 }} &  {69.15\tiny{$\pm$1.36 }}  \\

RandMSAugment $lr=0.03$  & \bf{14.48\tiny{$\pm$0.27}} & \bf{34.61\tiny{$\pm$0.22}} & {52.37\tiny{$\pm$0.09}} & \bf{69.96\tiny{$\pm$0.27}}\\

\bottomrule
    \end{tabular}
      \caption{Comparison between the Top-1 accuracies (\%, $\uparrow$) obtained with using a learning rate of 0.1 and 0.03 to train Tiny-Imagenet. $lr=0.03$ (second row) does better overall, even while the difference between the two cases is mostly small and under 1\%. }
\label{tab:lr}   
\end{table*}

\subsection{More RandMSAugment Examples}

\begin{figure*}[t]
\centering
\scriptsize
\begin{tabular}{@{}cccccc@{}}
\includegraphics[scale=0.73]{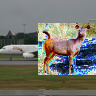} &
\includegraphics[scale=1.1]{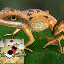} &
\includegraphics[scale=0.73]{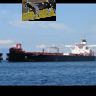} &
\includegraphics[scale=1.1]{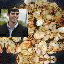} &
\includegraphics[scale=0.73]{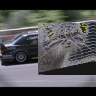} &
\includegraphics[scale=1.1]{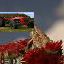} \\
\multicolumn{6}{c}{\protect\FFeatTransform, \protect\FResizePaste}\\

\includegraphics[scale=0.73]{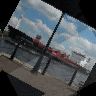} &
\includegraphics[scale=0.73]{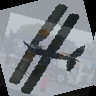} &
\includegraphics[scale=0.73]{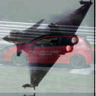} &
\includegraphics[scale=0.73]{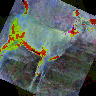} &
\includegraphics[scale=0.73]{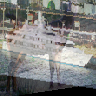} &
\includegraphics[scale=0.73]{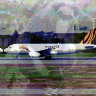} \\
\multicolumn{6}{c}{\protect\FFeatTransform, \protect\FInterpolate}\\

\includegraphics[scale=0.73]{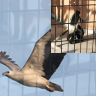} &
\includegraphics[scale=1.1]{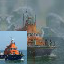} &
\includegraphics[scale=1.1]{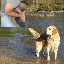} &
\includegraphics[scale=0.73]{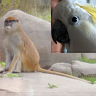} &
\includegraphics[scale=1.1]{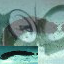} &
\includegraphics[scale=1.1]{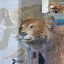}\\
\multicolumn{6}{c}{\protect\FResizePaste, \protect\FInterpolate}\\

\end{tabular}
\caption{More examples from RandMSAugment.}
\label{fig:samples}
\end{figure*}


\end{document}